\documentclass[10pt,letterpaper]{article}
\usepackage{hologene_preprint}
\date{}

\usepackage[utf8]{inputenc}
\usepackage[T1]{fontenc}
\usepackage{microtype}

\usepackage{amsmath}
\usepackage{amssymb}
\usepackage{mathtools}
\usepackage{amsthm}
\usepackage{amsfonts}
\usepackage{nicefrac}

\usepackage{graphicx}
\usepackage{subcaption}
\usepackage{booktabs}
\usepackage{multirow}
\usepackage{adjustbox}
\usepackage{float}

\usepackage{xcolor,colortbl}
\usepackage{enumitem}
\usepackage{xspace}
\usepackage{tikz}
\usetikzlibrary{positioning}

\usepackage[hyperfootnotes=false]{hyperref}
\usepackage{url}
\hypersetup{
  hidelinks,
  pdftitle={HoloGene: Learning to Lift Sliced Spatial Transcriptomics to Holistic 3D Gene Fields},
  pdfauthor={Yumin Zheng, Qingtian Zhu, Ziyan Zhu, Kailu Song, Yinqiang Zheng, Jun Ding}
}
\setcitestyle{authoryear,round}

\usepackage[capitalize,noabbrev]{cleveref}

\makeatletter
\DeclareRobustCommand\onedot{\futurelet\@let@token\@onedot}
\def\@onedot{\ifx\@let@token.\else.\null\fi\xspace}

\def\eg{\emph{e.g}\onedot}

\def\etal{\emph{et al}\onedot}
\makeatother

\newcommand{\ourmethod}{HoloGene\xspace}

\title{\ourmethod: Learning to Lift Sliced Spatial Transcriptomics
  to Holistic 3D Gene Fields}

\author{
  \begin{tabular}{c@{\hspace{2.4em}}c@{\hspace{2.4em}}c}
    \textbf{Yumin Zheng}\rlap{$^{1,2,*}$} &
    \textbf{Qingtian Zhu}\rlap{$^{3,*}$} &
    \textbf{Ziyan Zhu}\rlap{$^{4}$} \\
    \textbf{Kailu Song}\rlap{$^{1,2}$} &
    \textbf{Yinqiang Zheng}\rlap{$^{3,\dagger}$} &
    \textbf{Jun Ding}\rlap{$^{1,2,5,\dagger}$}
  \end{tabular} \\[0.3em]
  \textsuperscript{1}McGill University \quad
  \textsuperscript{2}MUHC Research Institute \quad
  \textsuperscript{3}The University of Tokyo \\
  \textsuperscript{4}Carnegie Mellon University \quad
  \textsuperscript{5}Mila -- Quebec AI Institute
}

\begin{document}

\maketitle
\begingroup\renewcommand{\thefootnote}{}\footnotetext{$^*$ Equal contribution. $^\dagger$ Corresponding authors: \texttt{yqzheng@ai.u-tokyo.ac.jp}, \texttt{jun.ding@mcgill.ca}.}\endgroup

\begin{abstract}
Spatial transcriptomics (ST) provides spatially resolved gene expression measurements, but most 3D ST datasets are acquired as independently processed 2D sections. Computational approaches span slice registration and integration, explicit or generative interpolation, and coordinate-field modeling. Within the last family, implicit neural representations (INRs) offer a continuous, coordinate-queryable parameterization, but standard isotropic coordinate encodings do not explicitly reflect that serial-section measurements are dense in-plane and sparse along the axial direction.
We present \textbf{\ourmethod}, an anisotropic conditional INR framework for estimating continuous 3D gene-expression fields from pre-registered ST sections. Its central contribution is an acquisition-aware factorization that couples dense in-plane synthesis with sparse axial evolution using established GAE, FiLM, pseudo-label supervision, and CORAL components. \ourmethod compresses high-dimensional expression profiles into a graph-autoencoder latent manifold, uses pseudo-label-derived signature conditioning to provide coarse intra-section context, and applies depth-dependent FiLM modulation to model axial variation. A signature-stratified correlation-alignment regularizer encourages cross-slice covariance alignment in latent features. Using held-out measured sections as cross-sectional references, \ourmethod improves reconstruction accuracy and biological preservation metrics compared with general-purpose INRs and adapted spatial-transcriptomics baselines. These results support anisotropic conditional modeling as a practical strategy for estimating 3D expression fields from discontinuous ST sections.
Code is available at \url{https://github.com/Szym29/HoloGene}.
\end{abstract}

\section{Introduction}
Spatial Transcriptomics (ST) transforms biological tissue analysis by mapping high-dimensional gene expression profiles directly onto their physical coordinates.
Advancing sequencing technologies are shifting the research focus from isolated 2D slices toward the construction of comprehensive 3D tissue atlases.
This transition is essential for studying biological processes that are inherently three-dimensional, such as organogenesis and the volumetric organization of complex tissues~\citep{chen2022spatiotemporal,wang2018three,zhang2023molecularly}.

Computational approaches to 3D ST include slice registration and integration~\citep{zeira2022alignment,qiu2024spatiotemporal}, explicit or generative interpolation~\citep{lin2025bridging,li2024high}, and Implicit Neural Representations (INRs)~\citep{zhu2025suica,luo2025stinr}.
Among these complementary families, INRs parameterize biological signals as continuous functions encoded within the weights of a neural network, typically a multilayer perceptron (MLP).
By learning a mapping from spatial coordinates to gene expression profiles, INRs support continuous coordinate queries without requiring a predefined output grid. This is an output parameterization choice rather than a substitute for registration, integration, or experimentally measured intermediate sections.

However, reconstructing such a continuous 3D gene expression field is hindered by a duality of structural bottlenecks: \textbf{physical anisotropy} and \textbf{data heterogeneity}.
The prevailing data acquisition workflow relies on mechanical sectioning, producing measurements that are dense and quasi-continuous in the XY-plane but sparse and discrete along the Z-axis.
The physical spacing between slices can range from a few micrometers to hundreds of micrometers, depending on the experimental design, whereas measurements within each slice are much denser.
This axial under-sampling creates a strong scale mismatch between the dense in-plane measurements and the sparse depth direction.
This severe scale disparity induces topological discontinuities, since biological structures may shift nonlinearly between adjacent sections.
Compounding this challenge are \textbf{inter-slice batch effects}: because slices are sequenced independently, technical variations across sections can masquerade as genuine biological signals.
Consequently, standard coordinate INRs that encode $(x,y,z)$ symmetrically~\citep{sitzmann2020implicit,tancik2020fourier} do not explicitly reflect the sampling asymmetry of serial ST, while models without a cross-slice alignment objective may retain slice-specific technical variation.

To address these acquisition-specific limitations, we propose \textbf{\ourmethod}, an anisotropic conditional INR framework tailored to serial-section ST, where dense intra-slice measurements coexist with sparse, independently acquired sections along the depth axis.
The central contribution is an acquisition-aware factorization that integrates established GAE, FiLM, pseudo-label supervision, and CORAL components around serial-section geometry. The synthesis network treats $(x,y)$ as coordinates of a dense in-plane field, while $z$ supplies layer-wise conditioning for axial variation.
A graph autoencoder provides compact latent regression targets. The Pioneer Signature Network distills Leiden partitions derived from training-section expression into a coordinate-predictable conditioning signal, and signature-stratified correlation alignment reduces covariance shifts among comparable coarse tissue states across slices.
This design produces coordinate-queryable estimates within the supplied pre-registered domain. Predictions at held-out measured sections can be compared with observed expression, whereas queries between measured sections are model-based interpolation estimates.
Under this evaluated protocol and across the included spatial technologies and tissue types, \ourmethod achieves the best result on most aggregate reconstruction and biological-preservation metrics among the compared methods.

\begin{figure}[t]
    \centering
    \includegraphics[width=\linewidth]{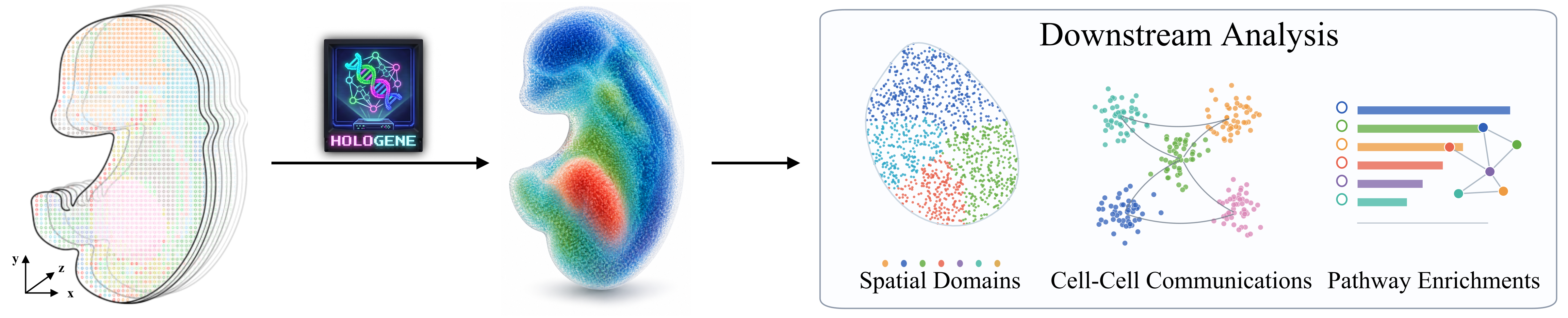}
    \caption{\ourmethod reconstructs holistic 3D gene fields from discontinuous ST sections.
By lifting sparsely sampled 2D slices into a continuous 3D field, \ourmethod enables volumetric downstream analysis, including spatial-domain recovery, cell--cell communication inference, and pathway enrichment.}
    \label{fig:teaser}
\vspace{-1.0em}
\end{figure}

\section{Related Works}

\subsection{3D Spatial Transcriptomics Reconstruction}
Reconstructing volumetric tissue architecture from anisotropic serial sections is a fundamental inverse problem in spatial biology.
Traditional approaches primarily rely on rigid or non-rigid registration algorithms to align adjacent slices.
\citet{zeira2022alignment} formulate slice alignment as an Optimal Transport (OT) problem in PASTE,
matching spots across layers based on transcriptional similarity and spatial proximity.
More recent frameworks shift toward graph-based integration and generative interpolation to bridge this dimensional gap.
For example, \citet{qiu2024spatiotemporal} employ Graph Neural Networks (GNNs)
for multi-slice alignment in Spateo,
whereas \citet{lin2025bridging} fill gaps along the Z-axis in SpatialZ
and construct dense 3D cell atlases by generating virtual intermediate slices.
These approaches provide complementary strengths in registration, integration, or virtual-slice generation. Coordinate-field models use a different output parameterization, supporting expression queries at locations within an aligned coordinate domain. This capability complements rather than replaces upstream registration or evidence from experimentally measured sections.
Furthermore, standard batch correction techniques (\eg, Harmony~\citep{korsunsky2019fast}, Seurat~\citep{hao2021integrated}) are typically decoupled from the spatial geometry but applied as preprocessing steps, potentially discarding spatially meaningful heterogeneity.

\subsection{Representation Learning for ST}
Owing to the high dimensionality and zero-inflated nature of ST data, representation learning plays a central role in extracting robust biological signals.
Discriminative models, such as SpaGCN~\citep{hu2021spagcn} and GraphST~\citep{long2023spatially}, are widely adopted, leveraging spatial neighborhood graphs to smooth gene expression and identify spatial domains.
Beyond discriminative models, generative approaches are increasingly used for tasks such as data imputation and super-resolution.
Notably, \citet{li2024high} employ generative modeling in STAGE
for conditional recovery of high-density gene expression from low-resolution measurements,
while \citet{zhu2025suica} use Graph Autoencoders (GAEs) in SUICA
to compress whole-transcriptome profiles into compact latent embeddings.
Our framework builds upon these representation-learning advances by adopting GAE-based compression to obtain compact latent regression targets.
STAGE, SUICA, and related approaches use different supervision schemes and output parameterizations for their respective reconstruction settings. Our focus is the complementary setting of a specimen-specific coordinate field whose in-plane synthesis and axial conditioning reflect the acquisition geometry of registered serial sections.

\subsection{Implicit Neural Representations in Biology}

INRs parameterize signals as continuous functions $\Phi(\mathbf{x})$, where $\mathbf{x}$ denotes coordinates. While this paradigm is widely used for 3D reconstruction in computer vision through coordinate-based MLPs, spatial biology introduces task-specific acquisition and technical-variation considerations.
More recently, this paradigm extends to biological data.
Works such as SUICA~\citep{zhu2025suica} and STINR~\citep{luo2025stinr} explore modeling gene expression fields with neural networks, supporting coordinate queries without a predefined output grid.
Directly extending a coordinate INR to multi-slice ST leaves two acquisition-specific issues implicit:
\begin{itemize}[noitemsep,topsep=0pt]
    \item \textbf{Isotropic Bias:} Symmetric $(x,y,z)$ coordinate encodings do not explicitly represent the asymmetry between the dense, quasi-continuous XY-plane and the sparse, discrete Z-axis.
    \item \textbf{Batch Sensitivity:} A coordinate regression objective alone does not explicitly distinguish independently acquired slice effects from biological variation.
\end{itemize}
\ourmethod addresses these issues through decoupled anisotropic conditioning and pseudo-label-stratified cross-slice alignment.

\begin{figure*}[t]
\centering
\includegraphics[width=\linewidth]{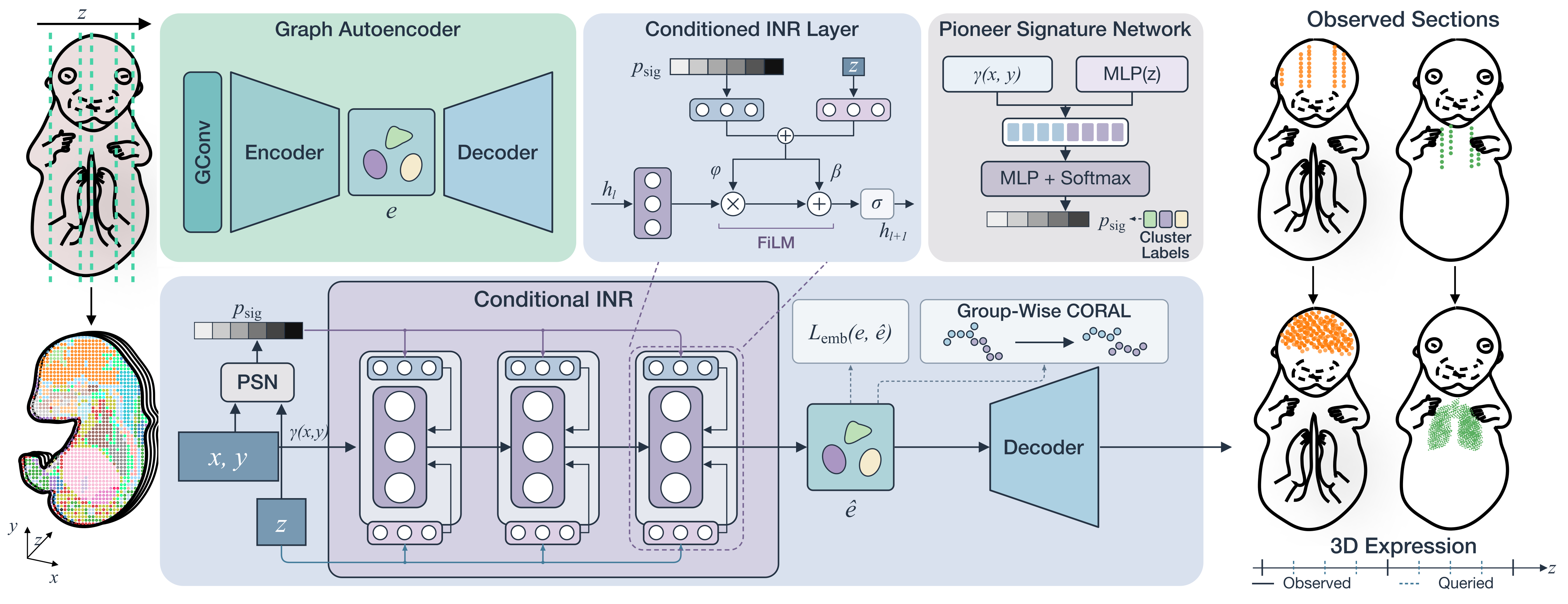}
\caption{Overview of \ourmethod. We first compress high-dimensional gene expression into latent embeddings via a graph autoencoder, then use PSN to distill training-expression partitions into coarse signature guidance, and synthesize a continuous 3D latent field using a decoupled, FiLM-conditioned INR with CORAL regularization to handle inter-slice anisotropy and regularize slice-specific representation variation.}
\label{fig:framework}
\vspace{-1.0em}
\end{figure*}

\section{Method}

\subsection{Overview}
Reconstructing a continuous 3D gene expression field from sliced ST data presents two fundamental challenges: the \textit{curse of dimensionality} inherent in whole-transcriptome profiles and the \textit{intrinsic data anisotropy} arising from sparse sampling along the Z-axis.

To address these, we propose \ourmethod. As illustrated in \cref{fig:framework}, the framework consists of three integrated components:
\begin{enumerate}[noitemsep,topsep=0pt]
    \item \textbf{Manifold Compression:} We project high-dimensional, sparse gene expression vectors into compact, topology-preserving latent embeddings using a GAE, following SUICA~\citep{zhu2025suica}.
    \item \textbf{Pioneer Signature Learning:} We predict training-expression-derived pseudo-label groups from 3D coordinates and use their probabilities to guide latent-field synthesis.
    \item \textbf{Anisotropic Latent Synthesis:} We synthesize a continuous 3D embedding field using a decoupled INR that conditions in-plane biological variation on depth-dependent axial context.
\end{enumerate}

\subsection{Manifold Compression}
Directly regressing high-dimensional, zero-inflated gene counts with an INR is computationally prohibitive and highly susceptible to noise overfitting.
Therefore, we adopt a manifold learning strategy inspired by SUICA~\citep{zhu2025suica} to compress the data into a dense latent representation.

We construct a spatial neighbor graph $\mathcal{G} = (\mathcal{V}, \mathcal{E})$ over the \emph{training slices only}, where $\mathcal{V}$ and $\mathcal{E}$ denote the node and edge sets, respectively, and each node $v_i$ represents a spot with raw gene expression $\mathbf{g}_i \in \mathbb{R}^G$.
Test slices are entirely excluded from graph construction and GCN message passing, ensuring no information leakage. 
A GCN encoder aggregates local neighborhood information to map $\mathbf{g}_i$ to a low-dimensional embedding $\mathbf{e}_i \in \mathbb{R}^D$ (typically $D=64$), preserving local tissue topology.
The network is pre-trained by minimizing the reconstruction loss. 
For subsequent INR training, we treat these learned embeddings $\{\mathbf{e}_i\}$ as latent regression targets, transforming the task from regressing sparse counts to fitting a smooth, denoised latent manifold that preserves the underlying tissue topology.
Note that similar to SUICA~\citep{zhu2025suica}, the training of GAE is performed as a separate step.

\subsection{Pioneer Signature Learning}

Before generating detailed tissue structures, we employ a Pioneer Signature Network (PSN) to map 3D coordinates to coarse tissue-state groups derived from training expression.
This module distills coordinate-predictable structure from expression-derived partitions into a conditioning signal for the subsequent synthesis phase; it does not discover biological states independently of expression.

PSN takes the full 3D coordinates $\mathbf{c} = (x, y, z)$ as input.
Concretely, the in-plane coordinates $(x,y)$ are first encoded via Gaussian/Fourier positional encoding to obtain spatial features,
while $z$ is passed through a small MLP to obtain depth features.
The two feature vectors are concatenated and fed into a two-layer MLP
that outputs biological-signature logits,
followed by softmax to produce the Signature Probability Distribution
$\mathbf{p}_{\text{sig}} \in \mathbb{R}^K$:
\begin{equation}
    \mathbf{p}_{\text{sig}} = \operatorname{softmax}\!\bigl(\mathrm{MLP}\bigl[\gamma(x,y)\,\|\,\mathrm{MLP}_z(z)\bigr]\bigr),
\end{equation}
where $\gamma(\cdot)$ denotes the positional encoding and $\|$ denotes concatenation.
Here, $K$ is determined by the Leiden algorithm~\citep{traag2019louvain}
applied to training-slice expression profiles
(using the default resolution of 1.0 in scanpy).
We use these Leiden-derived pseudo-labels for PSN supervision across all
datasets and do not use dataset-provided cell-type annotations as supervision
targets, avoiding label leakage from curated biological annotations.
We use the standard Cross-Entropy Loss ($\mathcal{L}_{\text{CE}}$) that enforces consistency with pseudo-labels for supervision.
This pseudo-label supervision provides coarse biological-state guidance for latent-field synthesis.

\subsection{Anisotropic Latent Synthesis}
The core of \ourmethod is a conditional INR designed to handle the anisotropy of ST data. 
Standard isotropic coordinate INRs do not explicitly encode the axial--lateral sampling imbalance of serial-section ST.
We propose a Decoupled Conditional Strategy that separates intra-slice manifold generation from inter-slice geometric evolution.
\citet{yin2023continuous} explore a related decoupling strategy in DINo,
where the temporal variable is modeled independently.

\subsubsection{Decoupled Inputs and Conditioning}

We restrict the main input of the synthesis network to intra-slice coordinates $(x, y)$. 
The Z-axis information is exclusively provided via conditioning, which encourages the network to learn a canonical 2D manifold that is continuously modulated to match the structure at depth $z$.

We utilize Feature-wise Linear Modulation (FiLM) to inject the decoupled conditions directly into the intra-slice spatial embeddings.
Specifically, we apply FiLM at every hidden layer. For a layer of width $H_l$, let $\mathbf{d}^{(l)}_z=\psi_l(z)\in\mathbb{R}^{2H_l}$ denote its learned depth condition, where $\psi_l$ is a layer-specific affine map. This condition is distinct from $\mathrm{MLP}_z(z)$ used by PSN. The depth condition and a learned embedding of $\mathbf{p}_{\mathrm{sig}}$ jointly produce the layer-wise scale and shift parameters. At the $l$-th layer, we modulate the hidden feature $\mathbf{h}_l$ to obtain $\mathbf{h}_{l+1}$:
\begin{equation}
    \mathbf{h}_{l+1} = \sigma\big(\boldsymbol{\phi}^{(l)} \odot \mathbf{h}_l + \boldsymbol{\beta}^{(l)}\big),
\end{equation}
where $\mathbf{h}_l$ denotes the hidden representation at layer $l$, $\sigma(\cdot)$ denotes the non-linear activation, and the layer-wise modulation parameters $\boldsymbol{\phi}^{(l)}$ (scaling) and $\boldsymbol{\beta}^{(l)}$ (shifting) combine the following two conditions:
\begin{itemize}[noitemsep,topsep=0pt]
    \item \textbf{Signature Condition ($\mathbf{p}_{\text{sig}}$):} The PSN probability distribution is mapped to a learned signature embedding that contributes to both scale and shift.
    \item \textbf{Z-Evolution Condition ($\mathbf{d}^{(l)}_z$):} The layer-specific encoding $\psi_l(z)$ contributes depth-dependent scale and shift, allowing the network to represent axial variation between measured slices.
\end{itemize}

\subsubsection{Synthesis Objective}
The synthesis network $\mathcal{S}_\theta$ is trained to regress the continuous latent manifold, which is later recovered to the raw gene space by the pre-trained decoder of GAE.
Since the target embeddings $\mathbf{e}_i$ lie in a continuous Euclidean space, we optimize the network parameters $\theta$ by minimizing the Mean Squared Error (MSE) over all valid spots $\Omega$:
\begin{equation}
    \mathcal{L}_{\text{emb}} = \frac{1}{|\Omega|} \sum_{i \in \Omega} \| \mathcal{S}_\theta(x_i, y_i \mid \mathbf{p}_{\text{sig}}^{(i)}, z_i) - \mathbf{e}_{i} \|^2_2,
\end{equation}
where $\mathbf{e}_{i}$ denotes the latent regression target for spot $i$ derived from the pre-trained GAE.
This objective forces the network to reconstruct a spatially continuous field that respects both the intra-slice topological structure (encoded in $\mathbf{e}_i$) and the inter-slice evolutionary consistency (enforced by the conditioning).

\begin{figure}[t]
\centering
\includegraphics[width=\linewidth]{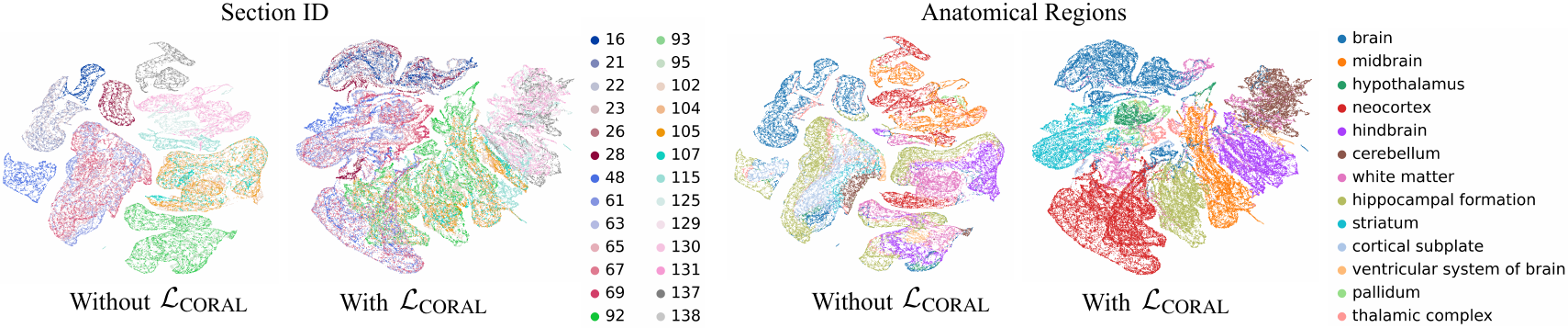}
\caption{Ablation on $\mathcal{L}_{\text{CORAL}}$: UMAP visualizations colored by section ID and anatomical regions on the whole mouse brain data~\citep{zhang2023molecularly}. Enabling $\mathcal{L}_{\text{CORAL}}$ leads to better cross-slice mixing and less visual separation by section ID.}
\label{fig:coral_ablation}
\vspace{-1.0em}
\end{figure}

A critical hurdle in 3D ST is the ``batch effect''---systematic technical variations between physical slices. 
If left unaddressed, the network may learn slice-specific artifacts as biological signals.
To mitigate this, we also introduce a Correlation Alignment (CORAL) Loss, inspired by Deep CORAL~\citep{sun2016deep}. 
This loss minimizes discrepancies between second-order statistics (covariance) of latent feature distributions for the same signature across different slices.
Let $\mathbf{C}_{k,s}\in\mathbb{R}^{D\times D}$ denote the covariance matrix
of $D$-dimensional latent features belonging to signature $k$ in slice $s$.
For each mini-batch, let $\mathcal{P}$ be the set of valid triples
$(k,s,s')$ for which pseudo-label group $k$ contains at least the configured
minimum number of spots in both slices $s$ and $s'$. The covariance component
of the batch-correction loss is
\begin{equation}
    \mathcal{L}_{\text{CORAL}}^{\text{cov}}
    = \frac{1}{|\mathcal{P}|D^2}
    \sum_{(k,s,s')\in\mathcal{P}}
    \left\|\mathbf{C}_{k,s}-\mathbf{C}_{k,s'}\right\|_F^2.
\end{equation}
In practice, we discard a signature group $k$ if it contains fewer than
$2\times\texttt{min\_n}$ spots across all slices in the mini-batch, and
exclude any slice-specific subset containing fewer than
$\texttt{min\_n}$ spots. Thus, only groups shared by at least two sufficiently
populated slices contribute to $\mathcal{P}$.
This conservative filtering prevents CORAL from forcing alignment
using poorly estimated covariance matrices from nearly empty subsets.
By minimizing $\mathcal{L}_{\text{CORAL}}^{\text{cov}}$, we reduce cross-slice
covariance discrepancies within each signature group and discourage
slice-specific variation in the latent features.
As illustrated in \cref{fig:coral_ablation}, enabling $\mathcal{L}_{\text{CORAL}}$
produces better cross-slice mixing in the embedding space and less visual
separation by slice identity. Together with the quantitative probes in
\cref{tab:supp_partition_free_coronal}, this supports the narrower conclusion
that CORAL reduces measured slice-identity leakage on the evaluated whole mouse
brain data.
Once trained, \ourmethod enables resolution-independent querying. 
For any queried coordinate in the pre-registered tissue domain, the network predicts an expression vector in $\mathbb{R}^G$. Predictions on held-out measured sections can be evaluated against observed expression, while queries between measured sections are model-based interpolation estimates.

\section{Experiments}

\subsection{Datasets \& Metrics}
We report the main results on two representative 3D spatial transcriptomics datasets:
MOSTA E16.5~\citep{chen2022spatiotemporal}, a volumetric mouse embryo atlas at embryonic day 16.5,
and a whole mouse brain dataset~\citep{zhang2023molecularly}.
Additional results on 3D Visual Cortex~\citep{wang2018three} and Breast Cancer~\citep{kuett2022three}
are provided in \cref{sec:supp_additional_results}.
For MOSTA E16.5, we use Spateo~\citep{qiu2024spatiotemporal} for spatial alignment in 2D across slices and place the slices in 3D space by directly adopting the z-axis coordinates from the original study~\citep{qiu2024spatiotemporal} to arrange slices in 3D;
the other datasets provide pre-registered 3D coordinates directly.

For quantitative evaluation, we follow SUICA~\citep{zhu2025suica} and report
mean squared error (MSE), cosine similarity (Cosine), Pearson and Spearman
correlations, and adjusted Rand index (ARI).
For expression reconstruction, MSE, Cosine, and Spearman are computed only at entries with ground-truth expression greater than zero (masked metrics, denoted by subscript $m$), whereas Pearson uses all entries without masking.
MSE/Cosine measure numerical fidelity, Pearson and Spearman measure correlation, and ARI compares clusters inferred from predicted expressions on held-out slices with matched dataset-provided cell-type or anatomical annotations. 
These annotations are used only for evaluation, while PSN is supervised by Leiden pseudo-labels from training-slice expressions.
We further report annotation-referenced biological preservation metrics, including DEG F1, Cell-Cell Communications (CCC) Pearson/Top-F1 (based on CellTalkDB~\citep{shao2021celltalkdb}), and pathway F1 using pathway enrichment resources; definitions are in \cref{sec:supp_biological_metrics}.
For all metrics, higher values indicate better performance, except for MSE, where lower values are preferred ($\downarrow$).

\subsection{Evaluation Protocol}
We adopt a specimen-specific, slice-wise measured-section holdout protocol to evaluate within-volume axial generalization in a pre-registered tissue coordinate domain. HoloGene estimates expression in the supplied coordinate system and does not jointly solve slice registration.

Concretely, we randomly split sections along the $z$-axis into 80\% for training and 20\% for testing. All spots from held-out measured sections are excluded from model training, GAE pre-training, graph construction, and GCN message passing.
The $z$-coordinates of test sections are never observed during training. 
We then evaluate whether the learned continuous estimator can recover the measured expression profiles, marker-gene patterns, annotation-consistent domains, and downstream biological readouts of these unseen physical sections.
The split is performed before every learned stage: the GAE graph and pretraining,
Leiden pseudo-label construction, PSN supervision, latent synthesis, and CORAL
regularization all use training sections only.
Held-out measured sections are used only for evaluation and visualization.

\begin{figure*}[t]
\vspace{-0.5em}
\centering
\includegraphics[width=0.93\linewidth]{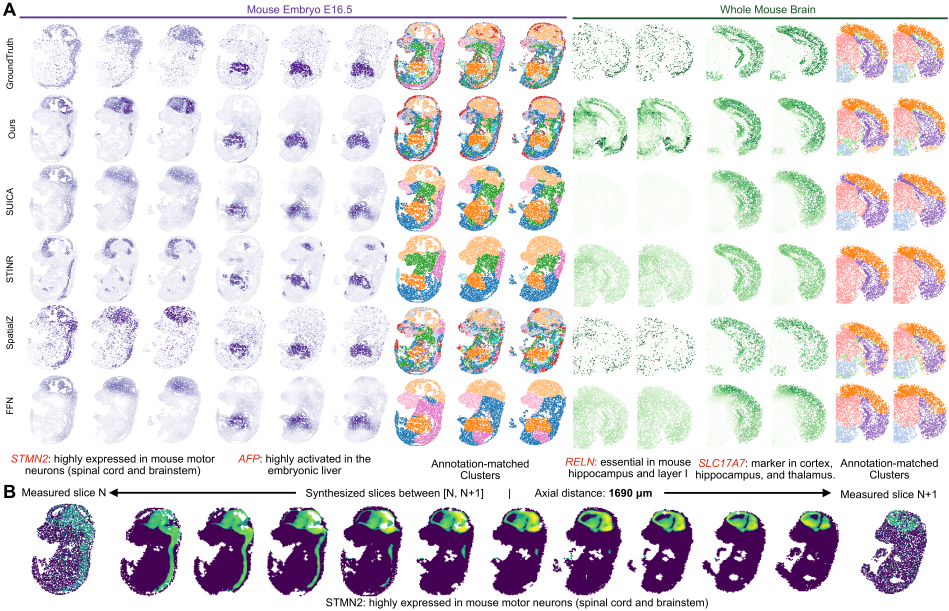}
\vspace{-0.3em}
\caption{Held-out section synthesis and inter-section querying.
A) Marker-gene and domain-map reconstruction on held-out measured sections.
Domain maps show ground-truth annotations in the first row and predicted-expression clusters mapped to annotation colors in the remaining rows. See full results in \cref{sec:supp_additional_results}.
B) Querying \ourmethod between measured sections provides model-based interpolation estimates with smooth STMN2 variation. See full results in \cref{sec:supp_evolution}.}
\label{fig:results}
\vspace{-1.2em}
\end{figure*}

\subsection{Baselines}
We compare \ourmethod against both general-purpose INRs and recent task-specific spatial transcriptomics methods.
Specifically, we include Fourier Feature Networks (FFN)~\citep{tancik2020fourier} and SIREN~\citep{sitzmann2020implicit} as representative INR baselines.
We also compare with recent spatial transcriptomics methods, including STAGE~\citep{li2024high}, SpatialZ~\citep{lin2025bridging}, STINR~\citep{luo2025stinr}, and SUICA~\citep{zhu2025suica}.
Note that STAGE and SUICA are originally designed for 2D spatial data.
For STAGE, we change the coordinate-matched latent bottleneck and its supervision target from $(x,y)$ to registered $(x,y,z)$ while retaining the original encoder, decoder, and objectives. For SUICA, we change the coordinate-MLP input from $\mathbb{R}^2$ to $\mathbb{R}^3$ and query registered $(x,y,z)$ while retaining its GAE, decoder, and objectives.
All methods receive the same aligned coordinates and use the same training and held-out physical sections and gene set; predictions are evaluated in the original expression space.
For INR-based baselines, we grid-search learning rate
$\in \{10^{-3}, 5 \times 10^{-4}, 10^{-4}\}$, batch size
$\in \{2048, 4096, 8192\}$, and latent dimension $\in \{32,64,128\}$;
STAGE follows its original training strategy. 
For SpatialZ, which is not a coordinate-INR baseline, we use the authors' default configuration from the released implementation.
In ablations, we use the same pretrained GAE latent targets and decoder to isolate the effects of anisotropic FiLM conditioning, PSN guidance, and CORAL alignment.

\subsection{Implementation Details}
We implement all methods in PyTorch and run experiments on a single NVIDIA A100 GPU. 
For graph-based components, the $k$-NN graph uses $k=5$ including self-loops. 
Baseline-specific training schedules follow the original implementations when available; for INR-based baselines, we use the grid search described above.
Runtime is reported in \cref{sec:supp_sensitivity_runtime}.

\subsection{Results}

\begin{table*}[t]
\centering
\caption{Quantitative benchmarking of virtual slice synthesis. Subscript $m$ denotes metrics evaluated only where ground-truth expression is positive. Bold/underline denote best/second-best based on unrounded values.}
\label{tab:results}
\small
\setlength{\tabcolsep}{4pt}
\begin{adjustbox}{max width=\linewidth}
\begin{tabular}{l|ccccc|ccccc}
\toprule
\multirow{2}{*}{Method} & \multicolumn{5}{c|}{\textbf{MOSTA E16.5}} & \multicolumn{5}{c}{\textbf{Whole Mouse Brain}} \\
& $\mathrm{MSE}_{m}\downarrow$ & $\mathrm{Cos}_{m}\uparrow$ & Pearson$\uparrow$ & $\mathrm{Spear}_{m}\uparrow$ & ARI$\uparrow$ & $\mathrm{MSE}_{m}\downarrow$ & $\mathrm{Cos}_{m}\uparrow$ & Pearson$\uparrow$ & $\mathrm{Spear}_{m}\uparrow$ & ARI$\uparrow$ \\
\midrule
FFN~\citep{tancik2020fourier} & 0.2588 & 0.7630 & 0.6525 & 0.5034 & 0.1009 & \underline{2.5616} & 0.7443 & 0.4414 & 0.4009 & \underline{0.4275} \\
SIREN~\citep{sitzmann2020implicit} & 0.2835 & 0.7373 & 0.6165 & 0.4695 & 0.0187 & 2.5778 & 0.7717 & \underline{0.4624} & \underline{0.4262} & 0.3837 \\
STAGE~\citep{li2024high} & 0.2987 & 0.7234 & 0.6655 & 0.5237 & 0.1213 & 2.9103 & 0.7356 & 0.4020 & 0.3662 & 0.3313 \\
SpatialZ~\citep{lin2025bridging} & 0.3145 & 0.7029 & 0.5315 & 0.4722 & 0.1188 & 3.3409 & 0.5279 & 0.2335 & 0.2070 & 0.0221 \\
STINR~\citep{luo2025stinr} & \underline{0.2332} & \underline{0.7971} & \underline{0.6986} & 0.5308 & \underline{0.1422} & 2.7411 & \underline{0.7782} & 0.4289 & 0.3895 & 0.2586 \\
SUICA-3D~\citep{zhu2025suica} & 0.2972 & 0.7233 & 0.6772 & \underline{0.5412} & 0.1393 & 2.7776 & \textbf{0.7803} & 0.4130 & 0.3839 & 0.3618 \\
\textbf{\ourmethod} (Ours) & \textbf{0.2303} & \textbf{0.8054} & \textbf{0.7146} & \textbf{0.5503} & \textbf{0.1637} & \textbf{2.4835} & 0.7748 & \textbf{0.4648} & \textbf{0.4315} & \textbf{0.4670} \\
\bottomrule
\end{tabular}
\end{adjustbox}
\end{table*}

\begin{table*}[t]
\centering
\caption{Downstream biological-readout preservation on MOSTA E16.5 and whole mouse brain. Bold/underline denote best/second-best based on unrounded values.}
\label{tab:bio_metrics}
\small
\setlength{\tabcolsep}{4pt}
\begin{adjustbox}{max width=\linewidth}
\begin{tabular}{l|cccc|cccc}
\toprule
\multirow{2}{*}{Method} & \multicolumn{4}{c|}{\textbf{MOSTA E16.5}} & \multicolumn{4}{c}{\textbf{Whole Mouse Brain}} \\
& Path. F1$\uparrow$ & CCC Pearson$\uparrow$ & CCC Top-F1$\uparrow$ & DEG F1$\uparrow$ & Path. F1$\uparrow$ & CCC Pearson$\uparrow$ & CCC Top-F1$\uparrow$ & DEG F1$\uparrow$ \\
\midrule
FFN~\citep{tancik2020fourier} & 0.5348 & 0.8654 & \underline{0.8892} & 0.4389 & 0.3462 & 0.9708 & 0.8800 & 0.8346 \\
SIREN~\citep{sitzmann2020implicit} & 0.3419 & 0.8444 & 0.8851 & 0.4056 & \underline{0.5538} & \underline{0.9767} & 0.8956 & 0.8615 \\
STAGE~\citep{li2024high} & 0.4068 & 0.8986 & 0.8880 & 0.2861 & 0.4242 & 0.8977 & 0.5444 & 0.6269 \\
SpatialZ~\citep{lin2025bridging} & \underline{0.6490} & 0.8806 & 0.8642 & \underline{0.5421} & 0.4935 & 0.9752 & \underline{0.8991} & \underline{0.8731} \\
STINR~\citep{luo2025stinr} & 0.5674 & 0.8786 & 0.8823 & 0.4306 & 0.4545 & 0.9618 & 0.8489 & 0.7192 \\
SUICA-3D~\citep{zhu2025suica} & 0.5163 & \underline{0.9122} & 0.8509 & 0.4167 & 0.3529 & 0.9245 & 0.8222 & 0.6692 \\
\textbf{\ourmethod} (Ours) & \textbf{0.7239} & \textbf{0.9342} & \textbf{0.9138} & \textbf{0.6278} & \textbf{0.5714} & \textbf{0.9804} & \textbf{0.9022} & \textbf{0.8808} \\
\bottomrule
\end{tabular}
\end{adjustbox}
\end{table*}

\begin{figure*}[t]
\vspace{-0.5em}
\centering
\includegraphics[width=0.93\linewidth]{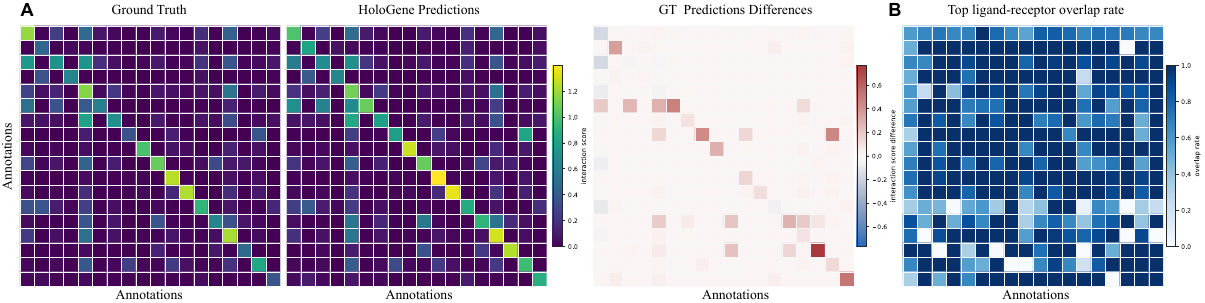}
\vspace{-0.1em}
\caption{\ourmethod preserves CCC patterns in MOSTA E16.5.
(A) Ground-truth and \ourmethod-predicted communication scores with their differences in the testing set.
(B) Overlap rate of top significant
  ligand-receptor interactions recovered by \ourmethod across annotation pairs. Darker blue indicates higher recovery.}
\label{fig:CCC}
\vspace{-0.8em}
\end{figure*}

\paragraph{Main benchmark on held-out measured slices.}
\cref{tab:results,tab:bio_metrics} and \cref{fig:results} summarize the main
performance comparison. \ourmethod achieves the lowest MSE and the highest Pearson, masked Spearman, and ARI scores on both MOSTA E16.5 and whole mouse brain data, as well as the highest masked Cosine on MOSTA E16.5. It also achieves the best scores on all four biological preservation metrics on both datasets; whole-brain masked Cosine is slightly below SUICA-3D and STINR.
On whole mouse brain, \ourmethod raises ARI from the strongest baseline score of 0.428 to 0.467, indicating better recovery of annotation-consistent tissue domains from held-out slices. 
Additional results on 3D Visual Cortex and Breast Cancer are reported in \cref{sec:supp_additional_results}, where \ourmethod remains competitive or best on representative reconstruction and biological preservation metrics.

\paragraph{Marker-level and domain-level reconstruction.}
The qualitative comparisons in \cref{fig:results} illustrate marker- and domain-level behavior across the two primary datasets. The selected markers
cover distinct anatomical programs. In MOSTA
E16.5, \textit{STMN2}~\citep{san2022loss} marks neuronal and motor-neuron-enriched structures, while
\textit{AFP}~\citep{kwon2006tg} highlights the embryonic liver region. In the whole mouse brain,
\textit{RELN}~\citep{vilchez2022specific,fatemi2011reelin} localizes to hippocampal and layer-I cortical structures, and
\textit{SLC17A7}~\citep{herzog2011vivo} marks glutamatergic excitatory neuronal populations in cortex,
hippocampus, and thalamus. 
The predictions preserve these region-specific patterns across multiple held-out z positions, and the annotation-matched domain maps show that expression-derived tissue organization remains spatially coherent after synthesis. Individual marker and section outcomes can differ across methods; our comparative conclusions are based on aggregate evaluation over all held-out measured sections. See full results in \cref{sec:supp_additional_results}.

\paragraph{Preservation of downstream biological readouts.}
Beyond marker-level reconstruction, we further evaluated whether downstream biological readouts remain recoverable from the synthesized expression fields.
In \cref{tab:bio_metrics}, \ourmethod achieves the best CCC Pearson and CCC Top-F1 on both datasets, indicating that ligand--receptor communication patterns
between tissue domains remain recoverable after virtual-slice synthesis. 
As an annotation-referenced case study on MOSTA E16.5 data, communication scores computed from predicted expressions closely match those from ground-truth expressions, with small residual differences and similar top sender-to-receiver connections (\cref{fig:CCC}, see full details in \cref{sec:ccc_details}).
These results suggest that \ourmethod preserves not only local marker-gene expression but also higher-order interaction structure inferred from spatially organized cell populations.

\paragraph{Continuous inter-section evolution.}
Beyond held-out measured-section reconstruction, \ourmethod can be queried at intermediate $z$-positions between measured sections. 
Because these locations do not have paired experimental measurements, we interpret the results as model-based interpolation estimates. 
\cref{fig:results}(B) shows that the estimated slices vary smoothly along the axial direction while preserving the \textit{STMN2}-enriched motor-neuron pattern observed in neighboring measured sections; the full two-marker visualization is provided in \cref{sec:supp_evolution}.

\paragraph{Ablations.}
\cref{tab:main_ablation_summary} analyzes the contribution of anisotropic depth conditioning, PSN guidance, and CORAL alignment under a shared setup where all variants use the same GAE latent targets and decoder. 
Comparing Isotropic XYZ with FiLM-only evaluates anisotropic decoupling: replacing direct $(x,y,z)$ input with $(x,y)$-based synthesis modulated by $z$ improves MSE and Pearson on both primary datasets.
In the FiLM-only versus FiLM+CORAL comparison, CORAL reduces whole-brain slice leakage from 0.1966 to 0.1770 and increases annotation retention from 0.4929 to 0.5102; label-independent diagnostics also show improved gene-level spatial autocorrelation and z-gradient preservation (\cref{tab:supp_partition_free_coronal}).
PSN contributes complementary signature-conditioned guidance, with metric-dependent benefits for biological preservation and annotation-consistent domain recovery.
Full \ourmethod does not maximize every metric or diagnostic independently; instead, it balances reconstruction, domain consistency, and biological preservation by combining FiLM-based depth conditioning, CORAL-based alignment, and PSN-based signature guidance.
Complete ablations, label-independent diagnostics, representation probes, split-ratio analysis, and runtime results are provided in 
\cref{sec:supp_ablation_extra,sec:supp_split_ratio,sec:supp_sensitivity_runtime,sec:supp_biological_metrics}.

\begin{table*}[t]
\centering
\caption{Component ablations. All variants use the same pretrained GAE. Bold/underline denote best/second-best based on unrounded values; full results are in \cref{sec:supp_ablation_extra,sec:supp_biological_metrics}.}
\label{tab:main_ablation_summary}
\small
\setlength{\tabcolsep}{4pt}
\textbf{(a) Reconstruction and annotation-consistent clustering}\par\smallskip
\begin{adjustbox}{max width=\linewidth}
\begin{tabular}{l|ccc|ccc}
\toprule
\multirow{2}{*}{Method} & \multicolumn{3}{c|}{\textbf{MOSTA E16.5}} & \multicolumn{3}{c}{\textbf{Whole Mouse Brain}} \\
& $\mathrm{MSE}_{m}\downarrow$ & Pearson$\uparrow$ & ARI$\uparrow$ & $\mathrm{MSE}_{m}\downarrow$ & Pearson$\uparrow$ & ARI$\uparrow$ \\
\midrule
Isotropic XYZ & 0.2413 & 0.6869 & 0.1310 & 2.5460 & 0.4446 & 0.4155 \\
FiLM-only & 0.2354 & 0.7132 & 0.1454 & 2.5041 & 0.4636 & 0.4231 \\
FiLM + PSN & 0.2308 & 0.7134 & 0.1440 & 2.4899 & 0.4636 & \underline{0.4398} \\
FiLM + CORAL & \textbf{0.2235} & \textbf{0.7154} & \underline{0.1569} & \textbf{2.4715} & \underline{0.4644} & 0.4354 \\
Full \ourmethod & \underline{0.2303} & \underline{0.7146} & \textbf{0.1637} & \underline{2.4835} & \textbf{0.4648} & \textbf{0.4670} \\
\bottomrule
\end{tabular}
\end{adjustbox}
\par\medskip
\textbf{(b) Biological preservation}\par\smallskip
\begin{adjustbox}{max width=\linewidth}
\begin{tabular}{l|ccc|ccc}
\toprule
\multirow{2}{*}{Method} & \multicolumn{3}{c|}{\textbf{MOSTA E16.5}} & \multicolumn{3}{c}{\textbf{Whole Mouse Brain}} \\
& Path. F1$\uparrow$ & CCC Pearson$\uparrow$ & DEG F1$\uparrow$ & Path. F1$\uparrow$ & CCC Pearson$\uparrow$ & DEG F1$\uparrow$ \\
\midrule
Isotropic XYZ & 0.5810 & 0.8937 & 0.4583 & 0.5000 & 0.9671 & 0.8538 \\
FiLM-only & 0.6667 & 0.9287 & 0.5944 & 0.4000 & 0.9792 & \underline{0.8692} \\
FiLM + PSN & 0.6853 & 0.9233 & \underline{0.6167} & \underline{0.5352} & \underline{0.9818} & 0.8654 \\
FiLM + CORAL & \underline{0.6854} & \underline{0.9290} & 0.6056 & 0.5231 & \textbf{0.9829} & \underline{0.8692} \\
Full \ourmethod & \textbf{0.7239} & \textbf{0.9342} & \textbf{0.6278} & \textbf{0.5714} & 0.9804 & \textbf{0.8808} \\
\bottomrule
\end{tabular}
\end{adjustbox}
\end{table*}

\section{Conclusion}

We presented \ourmethod, a manifold-aware conditional INR framework for estimating continuous 3D expression fields from discontinuous spatial transcriptomics sections.
By combining GAE-based latent compression, FiLM-based depth conditioning, PSN-based signature guidance, and CORAL-based cross-slice alignment, \ourmethod addresses the anisotropic sampling and slice-specific variation inherent to serial-section ST data. 
Under a slice-wise measured-section holdout protocol, \ourmethod achieves strong reconstruction and biological-preservation performance across multiple 3D ST datasets, while ablations show that the proposed components provide complementary benefits for depth-aware synthesis, cross-slice alignment, and biological structure preservation.

\paragraph{Limitations.}
\ourmethod estimates expression over a pre-registered tissue coordinate domain rather than directly measuring a dense 3D transcriptome. 
Intermediate queries between measured sections should therefore be interpreted as model-based estimates, and downstream biological conclusions require independent experimental validation.
The current model produces point estimates without uncertainty quantification and requires pre-registered coordinates; the stress tests show performance degradation under larger residual misalignment and very sparse axial coverage.
Finally, PSN supervision relies on Leiden-derived pseudo-labels. Reconstruction is stable over the evaluated resolution and label-corruption ranges, while annotation-referenced domain recovery is more sensitive; behavior beyond these ranges remains uncharacterized.

\section*{Acknowledgments}
This work was supported by AMED [JP23tm0524003]; JSPS KAKENHI [24KK0209]; JST-Mirai Program [JPMJMI23G1]; the Canadian Institutes of Health Research (CIHR) [PJT-180505]; the Fonds de recherche du Qu\'{e}bec - Sant\'{e} (FRQS) [295298, 295299, 366764]; the Natural Sciences and Engineering Research Council of Canada (NSERC) [RGPIN2022-04399]; and the Meakins-Christie Chair in Respiratory Research.
Funded by the European Union (AlRls, Grant Agreement No. 101289094).
The authors declare no competing interests.

\paragraph{Use of LLM-based tools.}
LLM-based tools assisted with manuscript preparation and the implementation, debugging, and execution of experimental code. They are not components of the proposed HoloGene model. The authors reviewed the resulting code and experimental outputs and take responsibility for the reported results.

\bibliographystyle{plainnat}
\bibliography{ref}

\appendix

\newpage
\raggedbottom

\section{Additional Results}
\label{sec:supp_additional_results}

\begin{figure}[H]
\centering
\includegraphics[width=\linewidth]{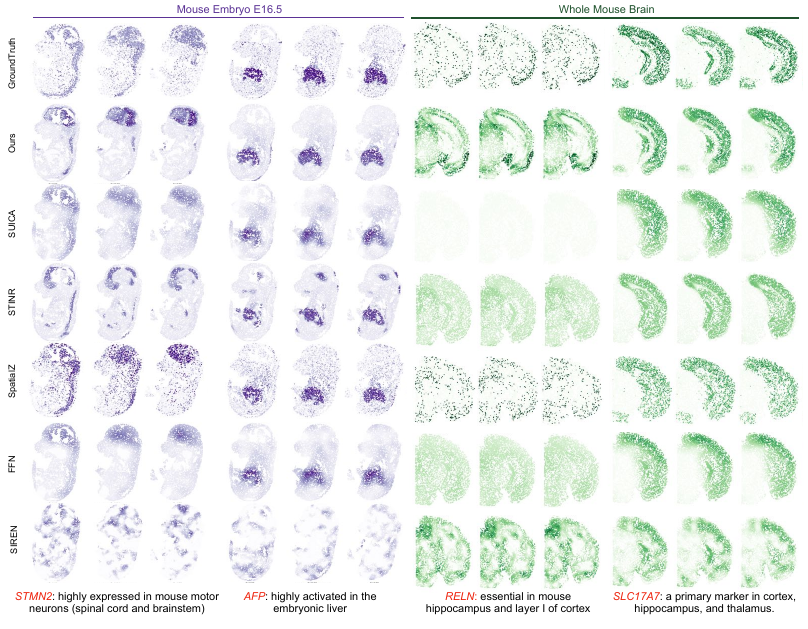}
\caption{Qualitative results of virtual slice synthesis on representative 3D samples. We visualize measured reference slices and the corresponding predictions from different methods; marker-level outcomes vary across genes and sections, while aggregate comparisons are reported in the main text.}
\label{fig:supp_full_results}
\end{figure}

\begin{figure}[H]
\centering
\includegraphics[width=\linewidth]{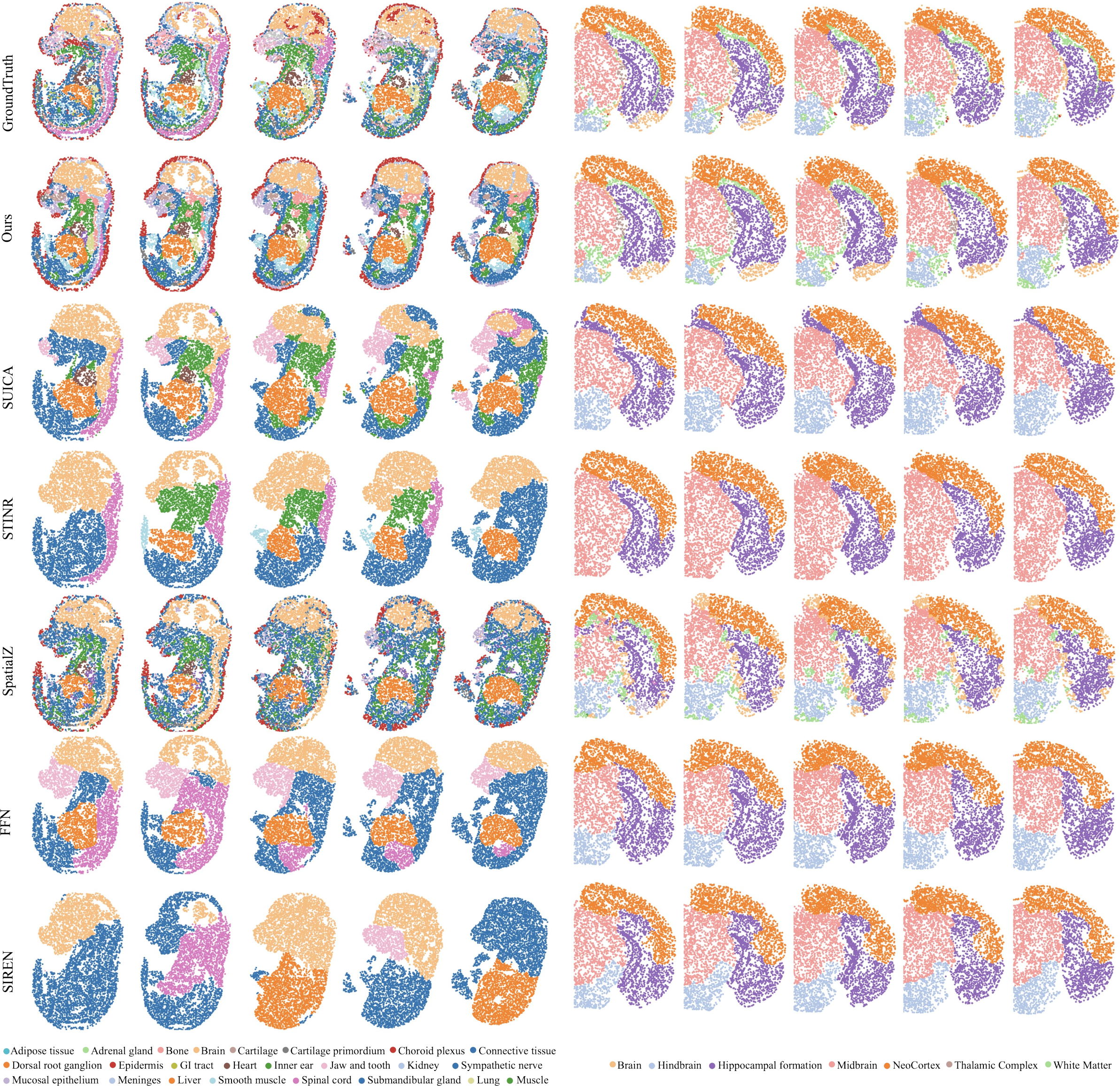}
\caption{Leiden-signature maps obtained from predicted gene expressions. \ourmethod yields more spatially coherent tissue regions with fewer isolated mislabeled spots compared with baselines.}
\label{fig:typing}
\end{figure}

We reported the full qualitative results of virtual slice synthesis on representative 3D samples in \cref{fig:supp_full_results}. The Leiden-signature maps obtained from predicted gene expressions can be found in \cref{fig:typing}.

We report additional benchmark results on 3D Visual Cortex~\citep{wang2018three}
and Breast Cancer~\citep{kuett2022three} in \cref{tab:supp_extra_reconstruction,tab:supp_visual_bio}.
For the breast cancer dataset, which contains a limited number of measured genes (25 genes), biological preservation metrics are not reported, as such metrics require broader transcriptomic coverage to be meaningful.
These datasets are used as supplementary robustness checks, while the main paper
focuses on MOSTA E16.5 and whole mouse brain.

\begin{table}[H]
\centering
\caption{Additional reconstruction results on 3D Visual Cortex and Breast Cancer. Subscript $m$ denotes masked metrics. Bold/underline denote best/second-best.}
\label{tab:supp_extra_reconstruction}
\small
\setlength{\tabcolsep}{4pt}
\textbf{(a) 3D Visual Cortex}\par\smallskip
\begin{adjustbox}{max width=\linewidth}
\begin{tabular}{lccccc}
\toprule
Method & $\mathrm{MSE}_{m}\downarrow$ & $\mathrm{Cos}_{m}\uparrow$ & Pearson$\uparrow$ & $\mathrm{Spear}_{m}\uparrow$ & ARI$\uparrow$ \\
\midrule
FFN~\citep{tancik2020fourier} & 7.7323 & 0.8126 & 0.4156 & 0.4195 & 0.1492 \\
SIREN~\citep{sitzmann2020implicit} & 6.1583 & 0.8413 & 0.4518 & 0.5402 & 0.0747 \\
STAGE~\citep{li2024high} & \underline{4.9072} & 0.8932 & 0.4989 & 0.6161 & 0.2134 \\
SpatialZ~\citep{lin2025bridging} & 8.3385 & 0.7488 & 0.3335 & 0.3554 & 0.0806 \\
STINR~\citep{luo2025stinr} & 4.9901 & \textbf{0.9197} & \underline{0.5421} & \textbf{0.6539} & 0.1770 \\
SUICA-3D~\citep{zhu2025suica} & 6.0817 & 0.8595 & 0.5186 & 0.5634 & \underline{0.3677} \\
\textbf{\ourmethod} (Ours) & \textbf{4.8994} & \underline{0.9050} & \textbf{0.5499} & \underline{0.6531} & \textbf{0.4198} \\
\bottomrule
\end{tabular}
\end{adjustbox}
\par\medskip
\textbf{(b) Breast Cancer}\par\smallskip
\begin{adjustbox}{max width=\linewidth}
\begin{tabular}{lcccc}
\toprule
Method & $\mathrm{MSE}_{m}\downarrow$ & $\mathrm{Cos}_{m}\uparrow$ & Pearson$\uparrow$ & $\mathrm{Spear}_{m}\uparrow$ \\
\midrule
FFN~\citep{tancik2020fourier} & 1.9190 & 0.9507 & 0.8179 & 0.8026 \\
SIREN~\citep{sitzmann2020implicit} & 0.8155 & 0.9793 & 0.9032 & 0.8754 \\
STAGE~\citep{li2024high} & 2.1255 & 0.9463 & 0.7632 & 0.7245 \\
SpatialZ~\citep{lin2025bridging} & 1.2619 & 0.9683 & 0.8557 & 0.8309 \\
STINR~\citep{luo2025stinr} & \underline{0.5906} & \underline{0.9851} & \underline{0.9203} & \underline{0.8958} \\
SUICA-3D~\citep{zhu2025suica} & 1.8693 & 0.9520 & 0.8059 & 0.7815 \\
\textbf{\ourmethod} (Ours) & \textbf{0.5134} & \textbf{0.9871} & \textbf{0.9317} & \textbf{0.9087} \\
\bottomrule
\end{tabular}
\end{adjustbox}
\end{table}

\begin{table}[H]
\centering
\caption{Additional biological preservation metrics on 3D Visual Cortex. Pathway F1 is omitted because the 155-gene panel is too small for stable pathway enrichment.}
\label{tab:supp_visual_bio}
\small
\setlength{\tabcolsep}{4pt}
\begin{adjustbox}{max width=\linewidth}
\begin{tabular}{lccc}
\toprule
Method & CCC Pearson$\uparrow$ & CCC Top-F1$\uparrow$ & DEG F1$\uparrow$ \\
\midrule
FFN~\citep{tancik2020fourier} & 0.8486 & 0.5000 & 0.2833 \\
SIREN~\citep{sitzmann2020implicit} & \underline{0.8928} & \underline{0.5263} & 0.4111 \\
STAGE~\citep{li2024high} & 0.7982 & 0.4342 & 0.2396 \\
SpatialZ~\citep{lin2025bridging} & 0.8783 & 0.4737 & 0.4401 \\
STINR~\citep{luo2025stinr} & 0.8215 & 0.4737 & 0.2563 \\
SUICA-3D~\citep{zhu2025suica} & 0.8906 & 0.4079 & \underline{0.5111} \\
\textbf{\ourmethod} (Ours) & \textbf{0.9002} & \textbf{0.5789} & \textbf{0.5905} \\
\bottomrule
\end{tabular}
\end{adjustbox}
\end{table}

\section{Cell Cell Communication reconstructions on the testing set}
\label{sec:ccc_details}
\begin{figure}[H]
\centering
\begin{tikzpicture}
  \node[anchor=south west,inner sep=0] (cccnetworks) at (0,0)
    {\includegraphics[width=\linewidth]{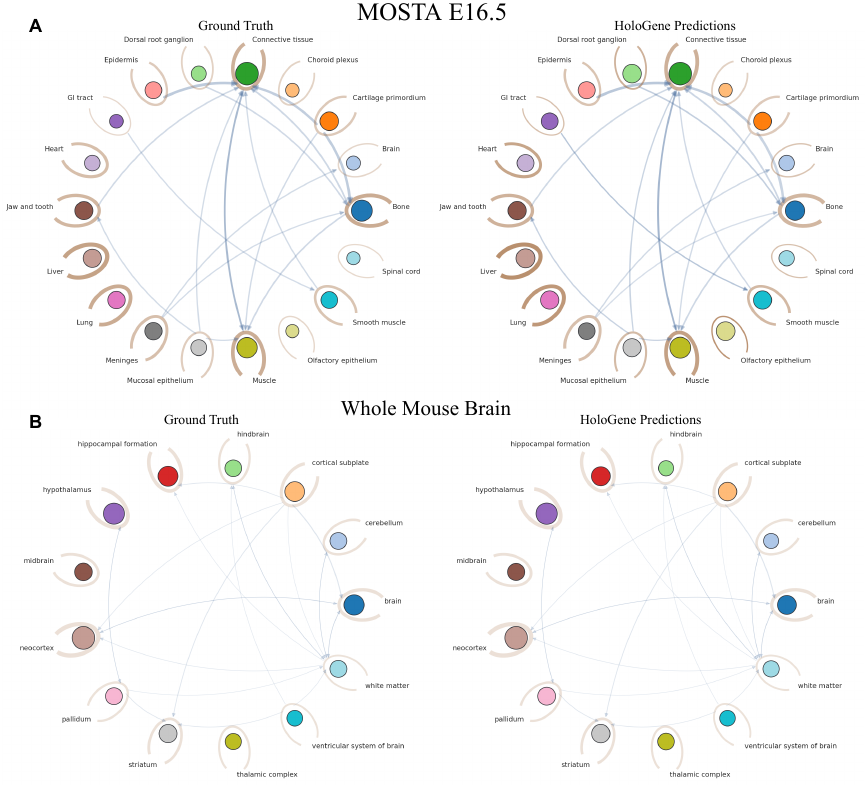}};
  \begin{scope}[x={(cccnetworks.south east)},y={(cccnetworks.north west)}]
    \fill[white] (0.35,0.955) rectangle (0.65,1.0);
    \node[font=\fontsize{10.5}{12}\selectfont] at (0.5,0.983) {MOSTA E16.5};
  \end{scope}
\end{tikzpicture}
\caption{Annotation-referenced CCC network preservation on (A) MOSTA E16.5 and (B) whole mouse brain.
Nodes denote anatomical annotations, and directed edges show top-scoring sender-to-receiver ligand--receptor communication connections.
Ground-truth and \ourmethod-predicted networks show similar high-confidence communication topology, indicating that virtual-slice synthesis preserves higher-order interaction structure.}
\label{fig:ccc_networks}
\end{figure}

\begin{figure}[H]
\centering
\includegraphics[width=\linewidth]{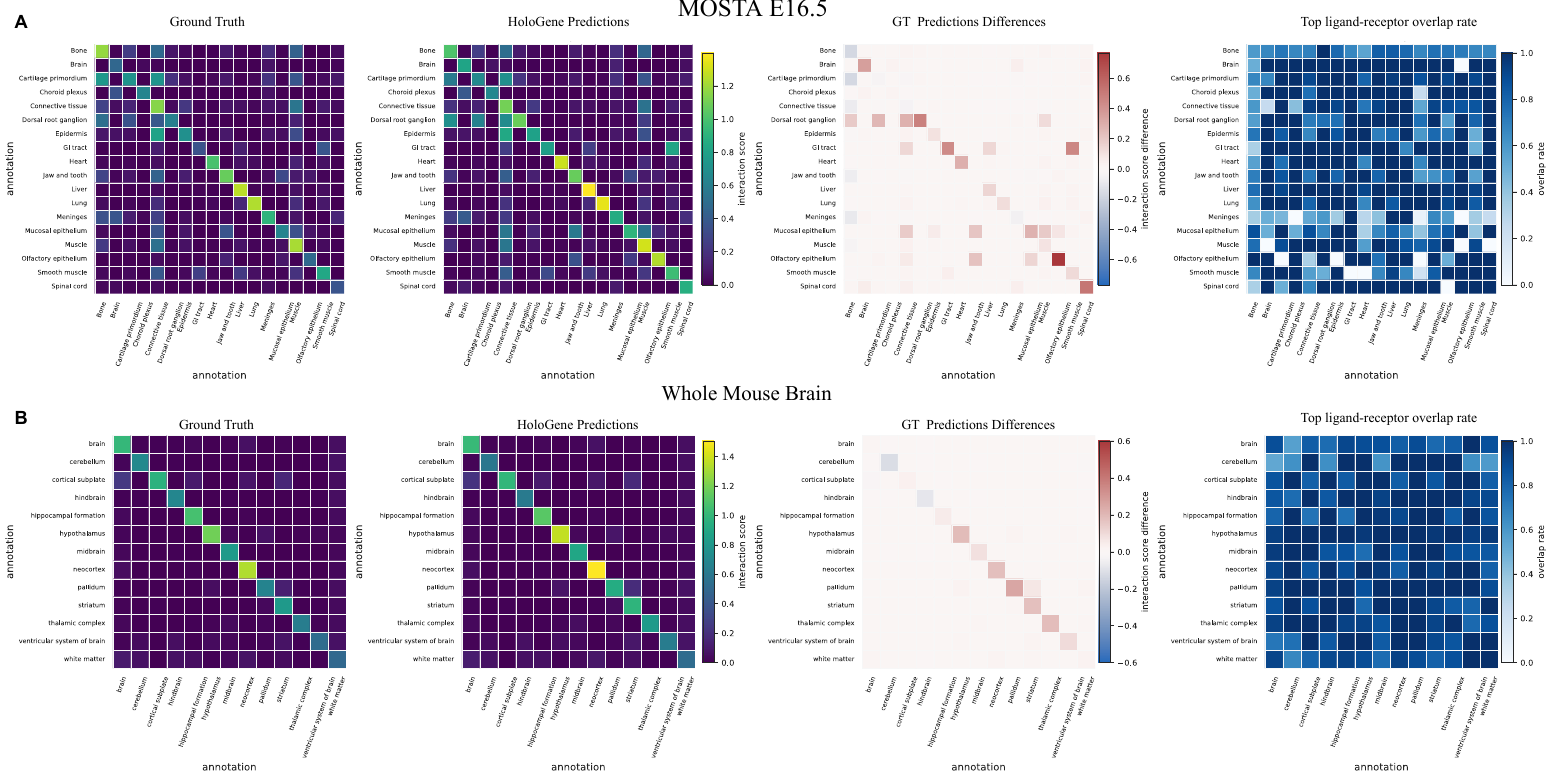}
\caption{Annotation-referenced cell--cell communication (CCC) preservation on the two primary datasets.
Rows and columns denote anatomical annotations used for post hoc sender--receiver grouping.
For each dataset, we compare ground-truth CCC interaction scores, HoloGene-predicted CCC interaction scores, prediction-minus-ground-truth residuals, and top ligand--receptor interaction overlap rates.
(A) MOSTA E16.5. (B) Whole mouse brain.
\ourmethod recovers similar label-pair communication structures to the ground truth, with small residual differences and high overlap of top ligand--receptor interactions across most annotation pairs.}
\label{fig:ccc_full}
\end{figure}

To further interpret the CCC preservation metrics, we provide annotation-referenced CCC visualizations in \cref{fig:ccc_full,fig:ccc_networks}. For each dataset, anatomical annotations are used only for post hoc sender--receiver grouping. We compute ligand--receptor communication scores from the ground-truth and HoloGene-predicted expression matrices using the same CCC procedure as in the main evaluation. \cref{fig:ccc_full} compares the full annotation-pair communication score matrices, their prediction-minus-ground-truth residuals, and the overlap rates of top ligand--receptor interactions. \cref{fig:ccc_networks} provides a complementary network view by visualizing the strongest sender-to-receiver communication edges. Across both MOSTA E16.5 and whole mouse brain, HoloGene largely preserves the dominant communication topology, with small residual differences and high overlap of top interactions, supporting the CCC Pearson and CCC Top-F1 improvements reported in \cref{tab:bio_metrics}.

\section{Visualization of Virtual Slice Evolution}
\label{sec:supp_evolution}
\begin{figure}[H]
\centering
\begin{subfigure}[t]{\linewidth}
  \centering
  \includegraphics[width=0.87\linewidth]{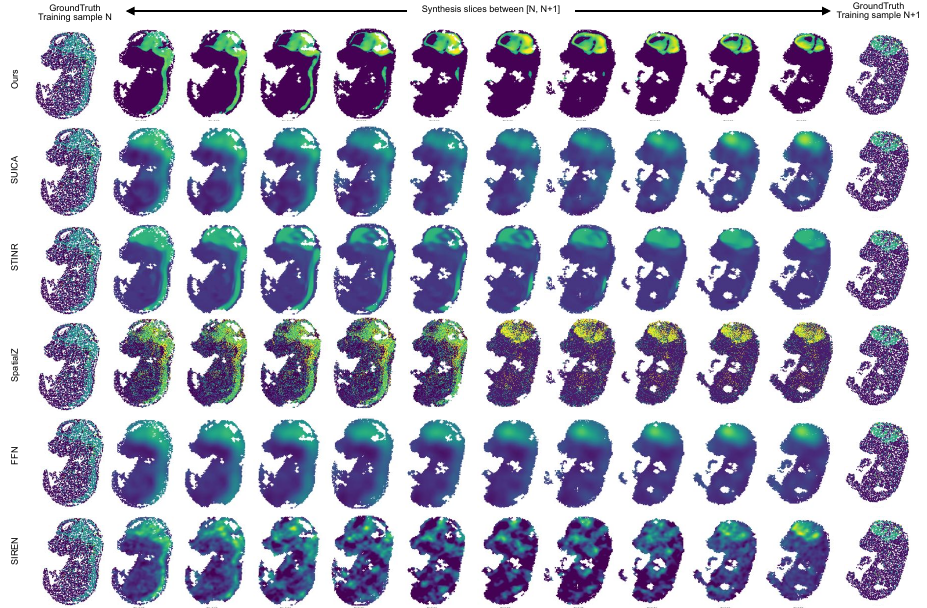}
  \caption{STMN2: highly expressed in mouse motor neurons (spinal cord and brainstem)}
\end{subfigure}

\vspace{0.5em}

\begin{subfigure}[t]{\linewidth}
  \centering
  \includegraphics[width=0.87\linewidth]{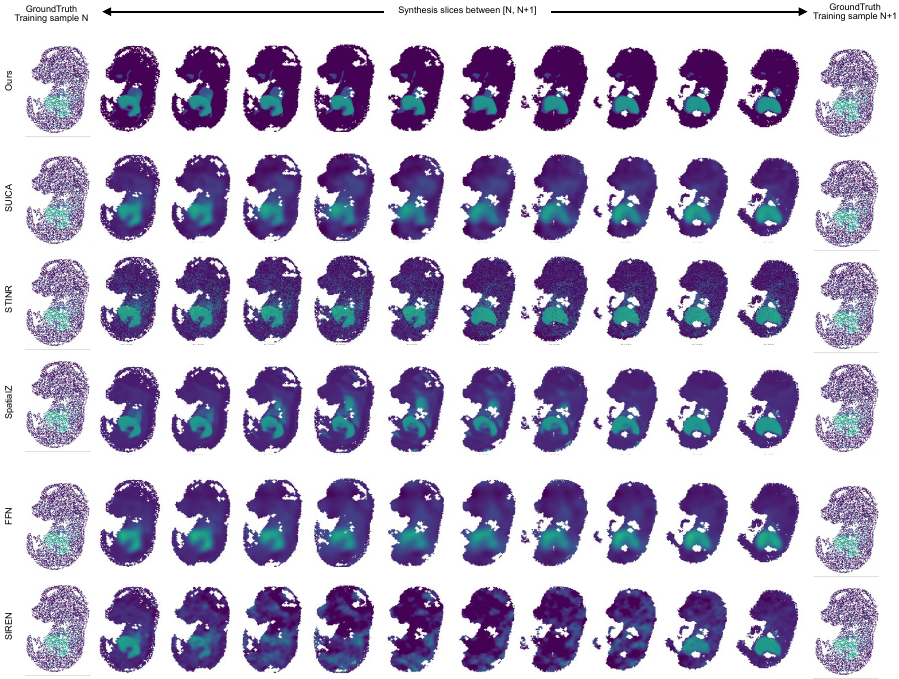}
  \caption{AFP: highly activated in the embryonic liver}
\end{subfigure}
\caption{Visualization of virtual-slice evolution for two gene markers. The leftmost and rightmost panels show measured slices, and the intermediate panels show synthesized virtual slices.}
\label{fig:evolution}
\end{figure}

We visualize the evolution of virtual slices between two measured slices for two gene markers in \cref{fig:evolution}. From left to right (or top to bottom), the leftmost and rightmost panels show the measured slices, while the intermediate panels show synthesized virtual slices, providing an intuitive assessment of the continuity and plausibility of the marker-expression patterns.

\clearpage

\clearpage
\section{Additional Ablations}
\label{sec:supp_ablation_extra}

We report unified ablations across reconstruction, clustering, and biological
preservation metrics in \cref{tab:supp_mosta_bio_ablation,tab:supp_coronal_bio_ablation,tab:supp_visual_bio_ablation}.
We use the following variants. Isotropic XYZ uses the same GAE compression and
decoder as \ourmethod, but removes FiLM conditioning, PSN, and CORAL, and
directly feeds $(x,y,z)$ coordinates to the INR. FiLM only uses the decoupled
FiLM-conditioned INR without PSN or CORAL. FiLM+PSN adds the PSN signature condition and supervision to the FiLM-only variant, whereas FiLM+CORAL adds only the CORAL loss, using the Leiden pseudo-label groups to define signature-conditioned covariance alignment when PSN is absent; evaluation metrics remain annotation-referenced as described in \cref{sec:supp_biological_metrics}.
Full \ourmethod uses FiLM, PSN, and CORAL jointly.

\begin{table}[H]
\centering
\caption{MOSTA E16.5 ablation study. Subscript $m$ denotes masked metrics. Higher is better except MSE.}
\label{tab:supp_mosta_bio_ablation}
\small
\setlength{\tabcolsep}{4pt}
\begin{adjustbox}{max width=\linewidth}
\begin{tabular}{lccccccccc}
\toprule
Model & Path. F1$\uparrow$ & CCC Pearson$\uparrow$ & CCC Top-F1$\uparrow$ & DEG F1$\uparrow$ & $\mathrm{MSE}_{m}\downarrow$ & $\mathrm{Cos}_{m}\uparrow$ & Pearson$\uparrow$ & $\mathrm{Spear}_{m}\uparrow$ & ARI$\uparrow$ \\
\midrule
Isotropic XYZ & 0.5810 & 0.8937 & 0.8865 & 0.4583 & 0.2413 & 0.7833 & 0.6869 & 0.5282 & 0.1310 \\
FiLM-only & 0.6667 & 0.9287 & 0.9056 & 0.5944 & 0.2354 & 0.7996 & 0.7132 & 0.5501 & 0.1454 \\
FiLM + PSN & 0.6853 & 0.9233 & 0.9070 & \underline{0.6167} & 0.2308 & 0.8040 & 0.7134 & 0.5495 & 0.1440 \\
FiLM + CORAL & \underline{0.6854} & \underline{0.9290} & \underline{0.9097} & 0.6056 & \textbf{0.2235} & \textbf{0.8067} & \textbf{0.7154} & \textbf{0.5512} & \underline{0.1569} \\
Full \ourmethod & \textbf{0.7239} & \textbf{0.9342} & \textbf{0.9138} & \textbf{0.6278} & \underline{0.2303} & \underline{0.8054} & \underline{0.7146} & \underline{0.5503} & \textbf{0.1637} \\
\bottomrule
\end{tabular}
\end{adjustbox}
\end{table}

\begin{table}[H]
\centering
\caption{Whole mouse brain ablation study. Subscript $m$ denotes masked metrics. Higher is better except MSE.}
\label{tab:supp_coronal_bio_ablation}
\small
\setlength{\tabcolsep}{4pt}
\begin{adjustbox}{max width=\linewidth}
\begin{tabular}{lccccccccc}
\toprule
Model & Path. F1$\uparrow$ & CCC Pearson$\uparrow$ & CCC Top-F1$\uparrow$ & DEG F1$\uparrow$ & $\mathrm{MSE}_{m}\downarrow$ & $\mathrm{Cos}_{m}\uparrow$ & Pearson$\uparrow$ & $\mathrm{Spear}_{m}\uparrow$ & ARI$\uparrow$ \\
\midrule
Isotropic XYZ & 0.5000 & 0.9671 & 0.8889 & 0.8538 & 2.5460 & 0.7519 & 0.4446 & 0.4125 & 0.4155 \\
FiLM-only & 0.4000 & 0.9792 & 0.8911 & \underline{0.8692} & 2.5041 & 0.7718 & 0.4636 & \underline{0.4309} & 0.4231 \\
FiLM + PSN & \underline{0.5352} & \underline{0.9818} & \textbf{0.9022} & 0.8654 & 2.4899 & 0.7720 & 0.4636 & 0.4305 & \underline{0.4398} \\
FiLM + CORAL & 0.5231 & \textbf{0.9829} & \underline{0.8933} & \underline{0.8692} & \textbf{2.4715} & \underline{0.7733} & \underline{0.4644} & 0.4297 & 0.4354 \\
Full \ourmethod & \textbf{0.5714} & 0.9804 & \textbf{0.9022} & \textbf{0.8808} & \underline{2.4835} & \textbf{0.7748} & \textbf{0.4648} & \textbf{0.4315} & \textbf{0.4670} \\
\bottomrule
\end{tabular}
\end{adjustbox}
\end{table}

\begin{table}[H]
\centering
\caption{3D Visual Cortex ablation study. Subscript $m$ denotes masked metrics. Higher is better except MSE. Pathway F1 is omitted because the 155-gene panel is too small for stable pathway enrichment.}
\label{tab:supp_visual_bio_ablation}
\small
\setlength{\tabcolsep}{4pt}
\begin{adjustbox}{max width=\linewidth}
\begin{tabular}{lcccccccc}
\toprule
Model & CCC Pearson$\uparrow$ & CCC Top-F1$\uparrow$ & DEG F1$\uparrow$ & $\mathrm{MSE}_{m}\downarrow$ & $\mathrm{Cos}_{m}\uparrow$ & Pearson$\uparrow$ & $\mathrm{Spear}_{m}\uparrow$ & ARI$\uparrow$ \\
\midrule
Isotropic XYZ & 0.8975 & 0.5263 & 0.4958 & 5.1350 & 0.8842 & 0.5133 & 0.6161 & 0.2986 \\
FiLM-only & \underline{0.8988} & 0.5132 & 0.5237 & 4.9825 & 0.9016 & 0.5496 & 0.6505 & 0.3819 \\
FiLM + PSN & 0.8985 & \underline{0.5658} & \underline{0.5627} & 4.9596 & \underline{0.9038} & \textbf{0.5541} & \textbf{0.6572} & 0.3269 \\
FiLM + CORAL & 0.8978 & 0.5263 & 0.5404 & \underline{4.9018} & 0.9000 & 0.5465 & 0.6491 & \textbf{0.4211} \\
Full \ourmethod & \textbf{0.9002} & \textbf{0.5789} & \textbf{0.5905} & \textbf{4.8994} & \textbf{0.9050} & \underline{0.5499} & \underline{0.6531} & \underline{0.4198} \\
\bottomrule
\end{tabular}
\end{adjustbox}
\end{table}

Overall, these ablations indicate a multi-objective trade-off rather than monotonic improvement from every added component. 
Compared with FiLM-only, FiLM+CORAL improves the measured spatial and representation diagnostics, whereas PSN improves selected signature-aware reconstruction and biological-preservation readouts. 
Full \ourmethod offers a favorable balance across the primary reconstruction, clustering, and downstream biological metrics, although it is not optimal for every diagnostic individually.

\section{Effect of Train/Test Split Ratio}
\label{sec:supp_split_ratio}

We study the performance of \ourmethod under different train/test split ratios with MOSTA E16.5~\citep{chen2022spatiotemporal}.
For each split, we train the model on the training slices and evaluate it on held-out test slices using the same protocol as in the main experiments.
\cref{tab:split-ratio} summarizes reconstruction, clustering, and biological
preservation metrics under each split.

\begin{table}[H]
\centering
\caption{MOSTA E16.5 performance under different train/test split ratios. The 8:2 split corresponds to the default setting. Subscript $m$ denotes metrics evaluated only where ground-truth expression is positive. Higher is better except MSE.}
\label{tab:split-ratio}
\small
\setlength{\tabcolsep}{4pt}
\begin{adjustbox}{max width=\linewidth}
\begin{tabular}{lccccccccc}
\toprule
Train:Test & $\mathrm{MSE}_{m}\downarrow$ & $\mathrm{Cos}_{m}\uparrow$ & Pearson$\uparrow$ & $\mathrm{Spear}_{m}\uparrow$ & ARI$\uparrow$ & Path. F1$\uparrow$ & CCC Pearson$\uparrow$ & CCC Top-F1$\uparrow$ & DEG F1$\uparrow$ \\
\midrule
8:2 & 0.2303 & 0.8054 & 0.7146 & 0.5503 & 0.1637 & 0.7239 & 0.9342 & 0.9138 & 0.6278 \\
6:4 & 0.2266 & 0.8039 & 0.7131 & 0.5485 & 0.0988 & 0.6388 & 0.9187 & 0.8996 & 0.5024 \\
5:5 & 0.3360 & 0.7310 & 0.6821 & 0.5246 & 0.0854 & 0.5422 & 0.9252 & 0.8651 & 0.3643 \\
4:6 & 0.2770 & 0.7490 & 0.6808 & 0.5312 & 0.0981 & 0.5996 & 0.9254 & 0.8894 & 0.4341 \\
2:8 & 0.2588 & 0.7782 & 0.6927 & 0.5384 & 0.0870 & 0.5170 & 0.9185 & 0.9036 & 0.3667 \\
\bottomrule
\end{tabular}
\end{adjustbox}
\end{table}

\section{Robustness to Axial Sampling Sparsity}
\label{sec:supp_axial_sampling}

The random train/test splits in \cref{tab:split-ratio} vary the number of
observed sections but do not isolate the axial spacing between them. We
therefore conduct a controlled-stride analysis in which every $k$-th measured
section is observed and all remaining sections are held out. This directly
controls the axial gap while retaining evaluation against measured expression
on held-out sections. On MOSTA E16.5, strides $k=2,3,4$ retain 7/13, 5/13,
and 4/13 sections, respectively. The MOSTA results in
\cref{tab:axial_stride} are averaged over three training seeds using the same
stride layout. The 150-section whole mouse brain volume
supports a wider range of strides, so we additionally evaluate
$k=2,4,6,8,10$, with three training seeds at each fixed stride layout.

\begin{table}[H]
\centering
\caption{Controlled axial-stride evaluation. Every $k$-th section is observed and the remaining measured sections are held out. Results are mean $\pm$ standard deviation over three training seeds at each fixed stride layout for both datasets. Higher is better except MSE.}
\label{tab:axial_stride}
\small
\setlength{\tabcolsep}{4pt}
\begin{adjustbox}{max width=\linewidth}
\begin{tabular}{llcccc}
\toprule
Dataset & Stride & $\mathrm{MSE}_{m}\downarrow$ & Pearson$\uparrow$ & ARI$\uparrow$ & Path. F1$\uparrow$ \\
\midrule
\multirow{3}{*}{MOSTA E16.5} & $k=2$ & 0.2344 $\pm$ 0.0018 & 0.7216 $\pm$ 0.0007 & 0.1225 $\pm$ 0.0014 & 0.6870 $\pm$ 0.0102 \\
 & $k=3$ & 0.2488 $\pm$ 0.0018 & 0.7160 $\pm$ 0.0009 & 0.1034 $\pm$ 0.0039 & 0.6916 $\pm$ 0.0121 \\
 & $k=4$ & 0.2250 $\pm$ 0.0019 & 0.7064 $\pm$ 0.0004 & 0.0976 $\pm$ 0.0039 & 0.6086 $\pm$ 0.0384 \\
\midrule
\multirow{5}{*}{Whole mouse brain} & $k=2$ & 2.5275 $\pm$ 0.0100 & 0.4657 $\pm$ 0.0004 & 0.3524 $\pm$ 0.0336 & 0.5483 $\pm$ 0.1100 \\
 & $k=4$ & 2.5003 $\pm$ 0.0385 & 0.4543 $\pm$ 0.0007 & 0.3363 $\pm$ 0.0110 & 0.5057 $\pm$ 0.0866 \\
 & $k=6$ & 2.5073 $\pm$ 0.0061 & 0.4476 $\pm$ 0.0001 & 0.3078 $\pm$ 0.0460 & 0.5387 $\pm$ 0.1810 \\
 & $k=8$ & 2.5592 $\pm$ 0.0163 & 0.4352 $\pm$ 0.0002 & 0.2679 $\pm$ 0.0146 & 0.4377 $\pm$ 0.0388 \\
 & $k=10$ & 2.5838 $\pm$ 0.0455 & 0.4280 $\pm$ 0.0026 & 0.2811 $\pm$ 0.0019 & 0.4983 $\pm$ 0.0126 \\
\bottomrule
\end{tabular}
\end{adjustbox}
\end{table}

On MOSTA E16.5, Pearson decreases from $0.722$ at $k=2$ to $0.706$ at
$k=4$, and ARI decreases from $0.1225$ to $0.0976$; MSE and Pathway F1 are
not monotonic across the tested strides. On whole mouse brain, Pearson
decreases from $0.466$ at $k=2$ to $0.428$ at $k=10$, while ARI decreases
from $0.352$ to $0.281$. The other metrics fluctuate across strides,
reflecting the different held-out sections as well as axial spacing.

To separate aggregate stride performance from position within a missing
interval, we reuse each trained stride model and group its held-out sections
by distance to the nearest observed section, measured in section intervals.
No model is retrained for an individual distance group. At $k=2$ and $k=3$
on MOSTA E16.5, all held-out sections have distance 1; $k=4$ is the first
nontrivial case and contains distances 1 and 2. For the longer whole mouse
brain volume, $k=6,8,10$ provide distances up to 5.

\begin{table}[H]
\centering
\caption{Within-stride performance stratified by distance to the nearest observed section. Results are mean $\pm$ standard deviation over the same three training seeds for both datasets. Higher is better except MSE.}
\label{tab:axial_distance}
\small
\setlength{\tabcolsep}{4pt}
\begin{adjustbox}{max width=\linewidth}
\begin{tabular}{llccccc}
\toprule
Dataset & Stride & Distance & $\mathrm{MSE}_{m}\downarrow$ & Pearson$\uparrow$ & ARI$\uparrow$ & Path. F1$\uparrow$ \\
\midrule
\multirow{2}{*}{MOSTA E16.5} & \multirow{2}{*}{$k=4$} & 1 & 0.2315 $\pm$ 0.0019 & 0.7084 $\pm$ 0.0004 & 0.1036 $\pm$ 0.0048 & 0.6351 $\pm$ 0.0278 \\
 &  & 2 & 0.2114 $\pm$ 0.0020 & 0.7024 $\pm$ 0.0006 & 0.0958 $\pm$ 0.0050 & 0.6790 $\pm$ 0.0171 \\
\midrule
\multirow{12}{*}{Whole mouse brain} & \multirow{3}{*}{$k=6$} & 1 & 2.5449 $\pm$ 0.0172 & 0.4497 $\pm$ 0.0004 & 0.3128 $\pm$ 0.0340 & 0.5363 $\pm$ 0.1363 \\
 &  & 2 & 2.4081 $\pm$ 0.0152 & 0.4526 $\pm$ 0.0004 & 0.3100 $\pm$ 0.0331 & 0.6237 $\pm$ 0.0538 \\
 &  & 3 & 2.6961 $\pm$ 0.0179 & 0.4329 $\pm$ 0.0010 & 0.3097 $\pm$ 0.0306 & 0.6254 $\pm$ 0.1478 \\
\cmidrule(lr){2-7}
 & \multirow{4}{*}{$k=8$} & 1 & 2.6804 $\pm$ 0.0204 & 0.4367 $\pm$ 0.0001 & 0.2719 $\pm$ 0.0196 & 0.6608 $\pm$ 0.0220 \\
 &  & 2 & 2.5129 $\pm$ 0.0208 & 0.4380 $\pm$ 0.0007 & 0.2817 $\pm$ 0.0204 & 0.7265 $\pm$ 0.1123 \\
 &  & 3 & 2.5870 $\pm$ 0.0224 & 0.4293 $\pm$ 0.0011 & 0.2804 $\pm$ 0.0200 & 0.4490 $\pm$ 0.0630 \\
 &  & 4 & 2.3339 $\pm$ 0.0210 & 0.4413 $\pm$ 0.0012 & 0.2783 $\pm$ 0.0185 & 0.3274 $\pm$ 0.0439 \\
\cmidrule(lr){2-7}
 & \multirow{5}{*}{$k=10$} & 1 & 2.5715 $\pm$ 0.0366 & 0.4458 $\pm$ 0.0015 & 0.2961 $\pm$ 0.0130 & 0.5739 $\pm$ 0.0609 \\
 &  & 2 & 2.4384 $\pm$ 0.0366 & 0.4407 $\pm$ 0.0016 & 0.2951 $\pm$ 0.0110 & 0.3842 $\pm$ 0.0730 \\
 &  & 3 & 2.6017 $\pm$ 0.0383 & 0.4175 $\pm$ 0.0018 & 0.2912 $\pm$ 0.0110 & 0.5640 $\pm$ 0.0423 \\
 &  & 4 & 2.5466 $\pm$ 0.0391 & 0.4199 $\pm$ 0.0023 & 0.2786 $\pm$ 0.0067 & 0.5640 $\pm$ 0.0147 \\
 &  & 5 & 2.8928 $\pm$ 0.0436 & 0.4004 $\pm$ 0.0026 & 0.2649 $\pm$ 0.0105 & 0.2219 $\pm$ 0.0085 \\
\bottomrule
\end{tabular}
\end{adjustbox}
\end{table}

For MOSTA E16.5 at $k=4$, the mid-gap distance 2 group has lower MSE
($0.211$ versus $0.232$) and higher Pathway F1 ($0.679$ versus $0.635$),
but lower Pearson and ARI than distance 1. Whole mouse brain also exhibits
non-monotonic distance effects at $k=6$ and $k=8$. The $k=10$ midpoint is
a clearer limitation: at distance 5, MSE reaches $2.893$, Pearson is
$0.400$, and Pathway F1 is $0.222$. Different distance groups contain
different anatomical sections, so distance alone does not determine performance.

Finally, we quantify variability due to section selection in the random-split
setting of \cref{tab:split-ratio}. At each held-out ratio from 20\% to 60\%,
we sample three independent section partitions. The results in
\cref{tab:axial_random_splits} therefore summarize partition variability
rather than training-seed variability.

\begin{table}[H]
\centering
\caption{Variability across random MOSTA E16.5 section partitions. Results are mean $\pm$ standard deviation over three independently sampled partitions at each held-out ratio. Higher is better except MSE.}
\label{tab:axial_random_splits}
\small
\setlength{\tabcolsep}{4pt}
\begin{adjustbox}{max width=\linewidth}
\begin{tabular}{lccccc}
\toprule
Held out & Observed & $\mathrm{MSE}_{m}\downarrow$ & Pearson$\uparrow$ & ARI$\uparrow$ & Path. F1$\uparrow$ \\
\midrule
20\% & 10 & 0.2403 $\pm$ 0.0377 & 0.7200 $\pm$ 0.0122 & 0.1548 $\pm$ 0.0142 & 0.7289 $\pm$ 0.0295 \\
30\% & 9 & 0.2379 $\pm$ 0.0331 & 0.7237 $\pm$ 0.0118 & 0.1346 $\pm$ 0.0139 & 0.7072 $\pm$ 0.0488 \\
40\% & 7 & 0.2470 $\pm$ 0.0201 & 0.7061 $\pm$ 0.0183 & 0.1026 $\pm$ 0.0086 & 0.6339 $\pm$ 0.0043 \\
50\% & 6 & 0.2813 $\pm$ 0.0474 & 0.6948 $\pm$ 0.0149 & 0.0995 $\pm$ 0.0144 & 0.5971 $\pm$ 0.0523 \\
60\% & 5 & 0.2539 $\pm$ 0.0211 & 0.7021 $\pm$ 0.0185 & 0.0941 $\pm$ 0.0003 & 0.6229 $\pm$ 0.0595 \\
\bottomrule
\end{tabular}
\end{adjustbox}
\end{table}

Together, the controlled-stride, distance-stratified, and repeated-partition
analyses distinguish the effect of axial gap from the number and identity of
observed sections. Pearson and domain recovery generally decline over the
broader stride range, but the individual metrics need not vary monotonically.
The most sparsely observed whole-brain midpoint illustrates a practical
limitation under very sparse axial coverage.

\section{Robustness to Registration and Pseudo-Label Perturbations}
\label{sec:supp_input_robustness}

\paragraph{Residual registration errors.}
Starting from the supplied alignment of MOSTA E16.5, we independently perturb
every training and held-out section while keeping expression values and the
section split fixed. All spots in one section receive the same rigid transform,
whereas perturbation directions and angles are sampled independently across
sections. Translation magnitudes follow the percentage settings used in the
perturbation runs, rotations are measured in degrees, and axial jitter is
expressed relative to the inter-section spacing. We then repeat the original
training and held-out evaluation protocol. \Cref{tab:registration_robustness}
reports mean $\pm$ standard deviation over three random seeds for all conditions, including the unperturbed reference.

\begin{table}[H]
\centering
\caption{Residual-registration stress test on MOSTA E16.5. All results, including the unperturbed reference, are mean $\pm$ standard deviation over three random seeds. Higher is better except MSE.}
\label{tab:registration_robustness}
\small
\setlength{\tabcolsep}{4pt}
\begin{adjustbox}{max width=\linewidth}
\begin{tabular}{lccc}
\toprule
Perturbation & $\mathrm{MSE}_{m}\downarrow$ & Pearson$\uparrow$ & ARI$\uparrow$ \\
\midrule
None & 0.2369 $\pm$ 0.0063 & 0.7115 $\pm$ 0.0003 & 0.1567 $\pm$ 0.0010 \\
Translation 1\% & 0.2340 $\pm$ 0.0016 & 0.7091 $\pm$ 0.0011 & 0.1515 $\pm$ 0.0045 \\
Translation 5\% & 0.2460 $\pm$ 0.0049 & 0.6877 $\pm$ 0.0081 & 0.1031 $\pm$ 0.0210 \\
Translation 10\% & 0.2556 $\pm$ 0.0088 & 0.6719 $\pm$ 0.0141 & 0.0780 $\pm$ 0.0078 \\
Rotation $1^\circ$ & 0.2318 $\pm$ 0.0012 & 0.7123 $\pm$ 0.0010 & 0.1553 $\pm$ 0.0033 \\
Rotation $5^\circ$ & 0.2362 $\pm$ 0.0030 & 0.7028 $\pm$ 0.0026 & 0.1404 $\pm$ 0.0092 \\
Rotation $10^\circ$ & 0.2417 $\pm$ 0.0043 & 0.6938 $\pm$ 0.0050 & 0.1185 $\pm$ 0.0096 \\
Axial jitter $0.25\times$ spacing & 0.2346 $\pm$ 0.0094 & 0.7109 $\pm$ 0.0014 & 0.1631 $\pm$ 0.0167 \\
Axial jitter $1\times$ spacing & 0.2391 $\pm$ 0.0121 & 0.7109 $\pm$ 0.0036 & 0.1582 $\pm$ 0.0138 \\
Translation 3\% + rotation $3^\circ$ & 0.2397 $\pm$ 0.0035 & 0.6973 $\pm$ 0.0053 & 0.1194 $\pm$ 0.0133 \\
Translation 10\% + rotation $10^\circ$ & 0.2575 $\pm$ 0.0044 & 0.6713 $\pm$ 0.0144 & 0.0737 $\pm$ 0.0152 \\
\bottomrule
\end{tabular}
\end{adjustbox}
\end{table}

Performance generally degrades as translation or rotation magnitude increases,
while small perturbations remain close to the unperturbed reference.
Pearson correlation is comparatively stable under axial jitter. This
experiment probes residual errors around a supplied registration and does not
establish robustness to completely unregistered or non-rigidly distorted data.

\paragraph{Pseudo-label resolution and corruption.}
The Leiden groups used to supervise PSN and stratify CORAL are constructed from
training-section expression only. They are distinct from the dataset-provided
annotations used to compute ARI after prediction. We first vary the Leiden
resolution used for these training-only groups. As shown in
\cref{tab:pseudolabel_resolution}, reconstruction metrics vary within a narrow
range, while annotation-referenced ARI is more sensitive to group granularity.

\begin{table}[H]
\centering
\caption{Sensitivity to the Leiden resolution used for training-only PSN and CORAL groups on MOSTA E16.5. Higher is better except MSE.}
\label{tab:pseudolabel_resolution}
\small
\setlength{\tabcolsep}{4pt}
\begin{adjustbox}{max width=\linewidth}
\begin{tabular}{lccc}
\toprule
Resolution & $\mathrm{MSE}_{m}\downarrow$ & Pearson$\uparrow$ & ARI$\uparrow$ \\
\midrule
0.5 & 0.2307 & 0.7133 & 0.1594 \\
0.8 & 0.2351 & 0.7129 & 0.1657 \\
1.0 & 0.2303 & 0.7146 & 0.1637 \\
1.2 & 0.2354 & 0.7106 & 0.1590 \\
1.5 & 0.2330 & 0.7150 & 0.1573 \\
\bottomrule
\end{tabular}
\end{adjustbox}
\end{table}

We next randomly shuffle 3\%, 5\%, or 10\% of training pseudo-label assignments
while fixing the section split and training seed. Nonzero corruption conditions
are averaged over independent corruption shuffles. The uncorrupted reference
and corruption runs do not share the same repetition source: the uncertainty
available for the former comes from training-seed variation, whereas the latter
varies corruption shuffles at a fixed seed. \Cref{tab:pseudolabel_corruption} reports mean $\pm$ standard deviation, with these distinct repetition sources identified explicitly.

\begin{table}[H]
\centering
\caption{Sensitivity to shuffled training pseudo-label assignments on MOSTA E16.5. Results are mean $\pm$ standard deviation. The uncorrupted reference varies training seeds; nonzero corruption conditions vary label shuffles at a fixed training seed. Higher is better except MSE.}
\label{tab:pseudolabel_corruption}
\small
\setlength{\tabcolsep}{4pt}
\begin{adjustbox}{max width=\linewidth}
\begin{tabular}{lccc}
\toprule
Corrupted labels & $\mathrm{MSE}_{m}\downarrow$ & Pearson$\uparrow$ & ARI$\uparrow$ \\
\midrule
0\% & 0.2278 $\pm$ 0.0032 & 0.7140 $\pm$ 0.0022 & 0.1691 $\pm$ 0.0049 \\
3\% & 0.2306 $\pm$ 0.0000 & 0.7140 $\pm$ 0.0005 & 0.1590 $\pm$ 0.0013 \\
5\% & 0.2337 $\pm$ 0.0014 & 0.7134 $\pm$ 0.0006 & 0.1529 $\pm$ 0.0051 \\
10\% & 0.2325 $\pm$ 0.0028 & 0.7136 $\pm$ 0.0013 & 0.1626 $\pm$ 0.0004 \\
\bottomrule
\end{tabular}
\end{adjustbox}
\end{table}

Across the evaluated resolution and corruption ranges, expression-level
reconstruction remains comparatively stable, while annotation-referenced
domain recovery varies more. These controlled tests do not imply robustness
beyond the evaluated pseudo-label settings.

\section{Repeated-Run Robustness}
\label{sec:supp_repeated_runs}

We separate two sources of variability on MOSTA E16.5. For training
stochasticity, every method uses one fixed section split and is trained with
five seeds. For section-selection variability, every method uses the same five
independently sampled train/held-out section partitions and one fixed training
seed. Thus panels (a) and (b) of \cref{tab:seed_split_robustness} should
not be interpreted as repeated estimates of the same randomness source.

\begin{table}[H]
\centering
\caption{Reconstruction and clustering robustness on MOSTA E16.5. Results are mean $\pm$ standard deviation over five training seeds at a fixed split or five section splits at a fixed training seed. Higher is better except MSE.}
\label{tab:seed_split_robustness}
\small
\setlength{\tabcolsep}{4pt}
\textbf{(a) Five training seeds, fixed section split}\par\smallskip
\begin{adjustbox}{max width=\linewidth}
\begin{tabular}{lccc}
\toprule
Method & $\mathrm{MSE}_{m}\downarrow$ & Pearson$\uparrow$ & ARI$\uparrow$ \\
\midrule
FFN & 0.3131 $\pm$ 0.0025 & 0.6583 $\pm$ 0.0134 & 0.1235 $\pm$ 0.0084 \\
SIREN & 0.2764 $\pm$ 0.0093 & 0.6657 $\pm$ 0.0041 & 0.1092 $\pm$ 0.0113 \\
STAGE & 0.3100 $\pm$ 0.0038 & 0.5788 $\pm$ 0.0106 & 0.1087 $\pm$ 0.0140 \\
SpatialZ & 0.3131 $\pm$ 0.0091 & 0.5308 $\pm$ 0.0021 & 0.1468 $\pm$ 0.0041 \\
STINR & \textbf{0.2246 $\boldsymbol{\pm}$ 0.0032} & 0.6950 $\pm$ 0.0112 & 0.1339 $\pm$ 0.0102 \\
SUICA-3D & 0.3016 $\pm$ 0.0016 & 0.6711 $\pm$ 0.0107 & 0.1352 $\pm$ 0.0023 \\
\ourmethod & 0.2333 $\pm$ 0.0067 & \textbf{0.7121 $\boldsymbol{\pm}$ 0.0008} & \textbf{0.1533 $\boldsymbol{\pm}$ 0.0065} \\
\bottomrule
\end{tabular}
\end{adjustbox}
\par\medskip
\textbf{(b) Five section splits, fixed training seed}\par\smallskip
\begin{adjustbox}{max width=\linewidth}
\begin{tabular}{lccc}
\toprule
Method & $\mathrm{MSE}_{m}\downarrow$ & Pearson$\uparrow$ & ARI$\uparrow$ \\
\midrule
FFN & 0.3070 $\pm$ 0.0093 & 0.6663 $\pm$ 0.0114 & 0.0989 $\pm$ 0.0323 \\
SIREN & 0.2777 $\pm$ 0.0204 & 0.6697 $\pm$ 0.0123 & 0.1096 $\pm$ 0.0145 \\
STAGE & 0.3037 $\pm$ 0.0145 & 0.5708 $\pm$ 0.0101 & 0.1141 $\pm$ 0.0119 \\
SpatialZ & 0.3243 $\pm$ 0.0096 & 0.5296 $\pm$ 0.0069 & 0.1287 $\pm$ 0.0210 \\
STINR & 0.2495 $\pm$ 0.0138 & 0.7086 $\pm$ 0.0085 & 0.1284 $\pm$ 0.0101 \\
SUICA-3D & 0.3294 $\pm$ 0.0145 & 0.6725 $\pm$ 0.0094 & 0.1381 $\pm$ 0.0041 \\
\ourmethod & \textbf{0.2361 $\boldsymbol{\pm}$ 0.0109} & \textbf{0.7191 $\boldsymbol{\pm}$ 0.0092} & \textbf{0.1547 $\boldsymbol{\pm}$ 0.0149} \\
\bottomrule
\end{tabular}
\end{adjustbox}
\end{table}

\begin{table}[H]
\centering
\caption{Biological-readout robustness over five training seeds at a fixed section split on MOSTA E16.5. Results are mean $\pm$ standard deviation.}
\label{tab:seed_bio_robustness}
\small
\setlength{\tabcolsep}{4pt}
\begin{adjustbox}{max width=\linewidth}
\begin{tabular}{lcccc}
\toprule
Method & Path. F1$\uparrow$ & CCC Pearson$\uparrow$ & CCC Top-F1$\uparrow$ & DEG F1$\uparrow$ \\
\midrule
FFN & 0.113 $\pm$ 0.019 & 0.871 $\pm$ 0.033 & 0.802 $\pm$ 0.074 & 0.341 $\pm$ 0.012 \\
SIREN & 0.404 $\pm$ 0.014 & 0.901 $\pm$ 0.009 & 0.688 $\pm$ 0.011 & 0.410 $\pm$ 0.031 \\
STAGE & 0.384 $\pm$ 0.052 & 0.885 $\pm$ 0.020 & 0.890 $\pm$ 0.012 & 0.280 $\pm$ 0.022 \\
SpatialZ & 0.673 $\pm$ 0.019 & 0.902 $\pm$ 0.013 & 0.889 $\pm$ 0.025 & 0.552 $\pm$ 0.012 \\
STINR & 0.656 $\pm$ 0.041 & 0.870 $\pm$ 0.016 & 0.868 $\pm$ 0.016 & 0.497 $\pm$ 0.008 \\
SUICA-3D & 0.509 $\pm$ 0.012 & 0.847 $\pm$ 0.004 & 0.849 $\pm$ 0.001 & 0.402 $\pm$ 0.010 \\
\ourmethod & \textbf{0.681 $\boldsymbol{\pm}$ 0.037} & \textbf{0.928 $\boldsymbol{\pm}$ 0.004} & \textbf{0.903 $\boldsymbol{\pm}$ 0.006} & \textbf{0.600 $\boldsymbol{\pm}$ 0.015} \\
\bottomrule
\end{tabular}
\end{adjustbox}
\end{table}

\begin{table}[H]
\centering
\caption{Biological-readout robustness over five section splits at a fixed training seed on MOSTA E16.5. Results are mean $\pm$ standard deviation.}
\label{tab:split_bio_robustness}
\small
\setlength{\tabcolsep}{4pt}
\begin{adjustbox}{max width=\linewidth}
\begin{tabular}{lcccc}
\toprule
Method & Path. F1$\uparrow$ & CCC Pearson$\uparrow$ & CCC Top-F1$\uparrow$ & DEG F1$\uparrow$ \\
\midrule
FFN & 0.327 $\pm$ 0.219 & 0.872 $\pm$ 0.041 & 0.854 $\pm$ 0.049 & 0.358 $\pm$ 0.013 \\
SIREN & 0.430 $\pm$ 0.047 & 0.915 $\pm$ 0.026 & 0.710 $\pm$ 0.041 & 0.433 $\pm$ 0.028 \\
STAGE & 0.382 $\pm$ 0.014 & 0.916 $\pm$ 0.034 & 0.905 $\pm$ 0.013 & 0.306 $\pm$ 0.019 \\
SpatialZ & 0.642 $\pm$ 0.030 & 0.896 $\pm$ 0.026 & 0.868 $\pm$ 0.016 & 0.576 $\pm$ 0.053 \\
STINR & 0.638 $\pm$ 0.030 & 0.909 $\pm$ 0.030 & 0.890 $\pm$ 0.012 & 0.459 $\pm$ 0.020 \\
SUICA-3D & 0.526 $\pm$ 0.006 & 0.869 $\pm$ 0.021 & 0.859 $\pm$ 0.020 & 0.407 $\pm$ 0.009 \\
\ourmethod & \textbf{0.703 $\boldsymbol{\pm}$ 0.042} & \textbf{0.948 $\boldsymbol{\pm}$ 0.017} & \textbf{0.919 $\boldsymbol{\pm}$ 0.009} & \textbf{0.588 $\boldsymbol{\pm}$ 0.030} \\
\bottomrule
\end{tabular}
\end{adjustbox}
\end{table}

Across both analyses, \ourmethod has the highest Pearson and ARI and the
strongest four biological readouts; STINR has the lowest MSE under fixed-split
training-seed variation, while \ourmethod has the lowest MSE across section
splits. These results characterize the two evaluated variability sources
separately rather than implying invariance to either source.

\section{Runtime}
\label{sec:supp_sensitivity_runtime}

On MOSTA E16.5 (downsampled to 10\% per slice, single A100),
training times (seconds) are:
FFN~1320, SIREN~1305, STAGE~542, SpatialZ~791,
STINR~2262, SUICA~2083, and \ourmethod~2040
for the PSN/CORAL/FiLM synthesis stage.
GAE pretraining takes 636s.
\ourmethod's main synthesis stage is comparable to STINR and SUICA,
and the additional GAE pretraining stage is a one-time cost for latent
manifold construction.

\section{Dataset Licenses and Availability}
\label{sec:dataset_licenses}

We use previously released datasets and cite the corresponding original
publications in the main paper.
The MOSTA E16.5 dataset is distributed under CC BY 4.0.
The whole mouse brain dataset is distributed under CC BY 4.0.
The Breast Cancer dataset from Kuett \etal is distributed under CC BY 4.0.
The 3D Visual Cortex dataset is publicly available, but we did not identify
an official explicit license statement.
We do not redistribute the original datasets as new assets in this submission.

\section{Biological Preservation Metrics}
\label{sec:supp_biological_metrics}

This section details the biological preservation readouts used in
\cref{tab:bio_metrics}. For all metrics, prediction-derived statistics are compared with the matched ground-truth expression matrix on held-out test cells.
Higher is better for all biological preservation metrics.
ARI uses dataset-provided cell-type or anatomical annotations as held-out evaluation targets. 
DEG F1, CCC metrics, and pathway F1 are also computed based on these dataset-provided annotations.
These annotations are used only for evaluation and visualization; they are not used for model training, PSN supervision, CORAL regularization, model selection, or hyperparameter tuning.

\paragraph{Pathway enrichment resources.}
Pathway enrichment is performed with the g:Profiler g:GOSt API using the mouse organism setting (\texttt{mmusculus})~\citep{kolberg2023gprofiler}. We use GO Biological Process~\citep{gene2026gene}, Reactome~\citep{2026reactome}, WikiPathways~\citep{agrawal2024wikipathways}, and KEGG~\citep{kanehisa2023kegg} with FDR correction at threshold 0.05. For each method and dataset, the background is the evaluated gene universe rather than the whole mouse genome.
We retain terms with adjusted $p<0.05$, term size at most 1000, intersection
size at least 15, and at most the top 20 enriched terms per label; labels are
retained only when the ground truth has at least three significant pathways.
Pathway F1 is reported for MOSTA E16.5 and whole mouse brain, but not for 3D Visual Cortex
because its 155-gene panel is too small for stable enrichment.

\paragraph{Ligand-receptor resource.}
Cell-cell communication (CCC) metrics use the mouse ligand-receptor table from
CellTalkDB v1.0~\citep{shao2021celltalkdb}. The source database contains 2033
mouse ligand-receptor pairs. For each dataset, we retain only pairs whose
ligand and receptor genes are both present in the evaluated gene panel.

\paragraph{Top20 DEG micro F1.}
For each dataset-provided annotation label $l$, we perform one-vs-rest differential expression analysis separately on the ground-truth expression matrix and the predicted expression matrix.
We retain the top 20 up-regulated marker genes per label,
denoted $G_l^{GT}$ and $G_l^{Pred}$. The micro-averaged precision and recall are
\begin{equation}
\text{Precision}_{micro}
= \frac{\sum_l |G_l^{Pred} \cap G_l^{GT}|}{\sum_l |G_l^{Pred}|},
\qquad
\text{Recall}_{micro}
= \frac{\sum_l |G_l^{Pred} \cap G_l^{GT}|}{\sum_l |G_l^{GT}|}.
\end{equation}
The reported score is the corresponding micro F1. This metric measures whether
the model preserves label-specific marker genes.

\paragraph{CCC interaction Pearson.}
For each dataset-provided annotation label $l$ and gene $g$, let $\mu_{l,g}^{M}$ be the mean expression for method $M \in \{GT, Pred\}$ among cells with label $l$. 
For sender annotation $s$, receiver annotation $r$, and ligand-receptor pair $p=(g_L,g_R)$, the communication
score is
\begin{equation}
S_{s,r,p}^{M}
= \mu_{s,g_L}^{M} \cdot \mu_{r,g_R}^{M} \cdot A_{s,r},
\end{equation}
where $A_{s,r}$ is the spatial adjacency weight estimated from the spatial
nearest-neighbor graph. CCC interaction Pearson is the Pearson correlation
between flattened predicted and ground-truth CCC score tensors, computed over
the union of interactions with nonzero predicted or ground-truth scores.

\paragraph{CCC top interaction F1.}
We also evaluate whether the strongest communication events are recovered. 
The candidate universe contains all tuples $u=(s,r,p)$, where $s$ and $r$ are dataset-provided annotation labels and $p$ is a retained ligand-receptor pair. 
Let $U$ be the set of candidates with nonzero predicted or ground-truth communication scores. We set
\begin{equation}
K = \min(5000,\ \max(25,\ \lceil 0.05 |U| \rceil)).
\end{equation}
Ground truth and prediction are ranked independently over the same candidate
universe, and the top-$K$ sets $T^{GT}_K$ and $T^{Pred}_K$ are compared by
set-overlap F1. The same rule for $K$ is used for all methods within a dataset.

\paragraph{Pathway micro F1.}
For each dataset-provided annotation label $l$, we compare the set of significant enriched pathway terms in ground truth and prediction:
\begin{equation}
P_l^{GT} =
\{(\text{direction}, \text{source}, \text{term id})\}_{GT,l},
\qquad
P_l^{Pred} =
\{(\text{direction}, \text{source}, \text{term id})\}_{Pred,l}.
\end{equation}
Micro precision and recall pool pathway overlaps across retained labels, and
the reported pathway score is the corresponding micro F1.

\paragraph{Label-independent spatial diagnostics.}
To complement the annotation-referenced biological preservation metrics above, we also report label-independent diagnostics on the whole mouse brain dataset in \cref{tab:supp_partition_free_coronal}.
Gene Moran Pearson computes Moran's I for each gene on the ground-truth and predicted expression fields and reports the Pearson correlation across genes, measuring whether spatial autocorrelation
patterns are preserved without using cluster labels.
Gene gradient-$z$ Pearson computes, for each gene, the Pearson correlation between expression and the $z$-coordinate, then correlates these per-gene scores between ground truth and
prediction.
Pathway mean Pearson computes per-pathway Pearson correlation of
cell-wise z-scored Reactome pathway scores between ground truth and prediction; pathway neighborhood Pearson applies the same comparison after 3D spatial neighborhood averaging with a $k=15$ nearest-neighbor graph.

The same comparison also includes two representation probes.
Slice leakage is the linear-probe accuracy for predicting slice identity within
annotation groups; lower values indicate less slice-identity information.
Annotation retention is the linear-probe accuracy for predicting anatomical
annotation; higher values indicate greater retained biological information.
These annotations are used only for post hoc probing, not to train the representations.

\begin{table}[H]
\centering
\caption{Spatial biological and representation diagnostics on whole mouse brain. Slice leakage is lower-is-better; the other metrics are higher-is-better. Bold denotes the better score.}
\label{tab:supp_partition_free_coronal}
\small
\setlength{\tabcolsep}{4pt}
\begin{adjustbox}{max width=\linewidth}
\begin{tabular}{lcccccc}
\toprule
Model & Moran Pear.$\uparrow$ & Gene Grad. $z\uparrow$ & Path. Mean$\uparrow$ & Path. Nbhd.$\uparrow$ & Slice Leak$\downarrow$ & Ann. Ret.$\uparrow$ \\
\midrule
FiLM-only & 0.6040 & 0.9220 & 0.2655 & 0.6227 & 0.1966 & 0.4929 \\
FiLM + CORAL & \textbf{0.6103} & \textbf{0.9326} & \textbf{0.2662} & \textbf{0.6244} & \textbf{0.1770} & \textbf{0.5102} \\
\bottomrule
\end{tabular}
\end{adjustbox}
\end{table}

\paragraph{Axial-specific preservation diagnostics.}
To test whether reduced slice leakage is accompanied by loss of axial biology,
we isolate CORAL in the PSN-conditioned whole-mouse-brain model: ``HoloGene
w/o CORAL'' uses FiLM+PSN, whereas full \ourmethod uses FiLM+PSN+CORAL.
Gene-z Spearman measures preservation of per-gene monotonic axial profiles. Within
each cell type, pathway axial direction agreement compares the direction of
pathway activation changes along $z$, pathway axial trajectory Pearson compares
the full axial trajectories, and pathway axial signal ratio measures retained
variance in the fitted linear axial component.

\begin{table}[H]
\centering
\caption{Axial-specific preservation diagnostics on whole mouse brain. All metrics are higher-is-better.}
\label{tab:supp_axial_preservation}
\small
\setlength{\tabcolsep}{4pt}
\begin{adjustbox}{max width=\linewidth}
\begin{tabular}{lcc}
\toprule
Diagnostic & HoloGene w/o CORAL & Full \ourmethod \\
\midrule
Gene-z Spearman & 0.9200 & \textbf{0.9479} \\
Pathway axial direction agreement & 0.8169 & \textbf{0.8251} \\
Pathway axial trajectory Pearson & 0.4629 & \textbf{0.4740} \\
Pathway axial signal ratio & 0.4893 & \textbf{0.5145} \\
\bottomrule
\end{tabular}
\end{adjustbox}
\end{table}

All four measured scores improve when CORAL is enabled in this comparison.
Separately, the FiLM-only versus FiLM+CORAL comparison in
\cref{tab:supp_partition_free_coronal} improves all four label-independent
spatial diagnostics, reduces slice leakage, and increases annotation retention.
These complementary comparisons support CORAL on the evaluated tissue without
implying a guarantee for other tissues or sampling regimes.

\end{document}